\documentclass{article}

\usepackage[preprint]{nips_2018}

\usepackage{natbib}

\usepackage[utf8]{inputenc} 
\usepackage[T1]{fontenc}    
\usepackage{hyperref}       
\usepackage{url}            
\usepackage{booktabs}       
\usepackage{amsfonts}       
\usepackage{nicefrac}       
\usepackage{microtype}      
\usepackage{graphicx}

\usepackage{placeins}

\usepackage{comment}
\usepackage{amsmath}
\usepackage{amssymb}
\usepackage{subcaption}

\usepackage{algorithm} 
\usepackage{algpseudocode}

\newcommand{\vect}[1]{\boldsymbol{#1}}
\renewcommand{\vec}[1]{\boldsymbol{#1}}
\newcommand{\piold}{\pi_{\textrm{old}}}
\newcommand{\pioldsa}{\pi_{\textrm{old}}(\vec a | \vec s)}
\newcommand{\pisa}{\pi(\vec a | \vec s)}

\newcommand{\Vpi}{V^{\pi}}

\newcommand{\Api}{A^{\pi}}

\newcommand{\Qpi}{Q^{\pi}}
\newcommand{\Qpiold}{Q^{\pi_{\textrm{old}}}}
\newcommand{\Qpioldsa}{Q^{\pi_{\textrm{old}}}(\vec s,\vec a)}
\newcommand{\Qpioldsai}{Q^{\pi_{\textrm{old}}}(\vec s_i,\vec a_i)}
\newcommand{\Qpioldsat}{Q^{\pi_{\textrm{old}}}(\vec s_t,\vec a_t)}
\newcommand{\Qpisa}{Q^{\pi}(\vec s,\vec a)}

\newcommand{\vecto}[0]{\textrm{vec}}

\newcommand{\Acronym}{COPOS}

\title{Compatible Natural Gradient Policy Search}

\author{Joni Pajarinen \\
  Intelligent Autonomous Systems, TU Darmstadt, Germany\\
  \texttt{pajarinen@ias.tu-darmstadt.de}
  \And
  Hong Linh Thai \\
  Intelligent Autonomous Systems, TU Darmstadt, Germany
  \And
  Riad Akrour \\
  Intelligent Autonomous Systems, TU Darmstadt, Germany
  \And
  Jan Peters \\
  Intelligent Autonomous Systems, TU Darmstadt, Germany\\
  MPI for Intelligent Systems, Tuebingen, Germany
  \And
  Gerhard Neumann \\
  Lincoln Center for Autonomous Systems, University of Lincoln, Lincoln, UK
}

\begin{document}

\maketitle

\begin{abstract}
Trust-region methods have yielded state-of-the-art results in policy
search. A common approach is to use KL-divergence to bound the region
of trust resulting in a natural gradient policy update. We show that
the natural gradient and trust region optimization are equivalent if
we use the natural parameterization of a standard exponential policy
distribution in combination with compatible value function
approximation. Moreover, we show that standard natural gradient
updates may reduce the entropy of the policy according to a wrong
schedule leading to premature convergence. To control entropy
reduction we introduce a new policy search method called compatible
policy search (COPOS) which bounds entropy loss. The experimental
results show that COPOS yields state-of-the-art results in challenging
continuous control tasks and in discrete partially observable tasks.
\end{abstract}

\section{Introduction}
The natural gradient \citep{amari98natural} is an integral part of many reinforcement
learning \citep{Kakade:2001,bagnell2003covariant,4863,geist10} and optimization \citep{wierstraCEC2008} algorithms. Due to the natural gradient,
gradient updates become invariant to affine transformations of the
parameter space and the natural gradient is also often used to define
a trust-region for the policy update. The trust-region is defined by a
bound of the Kullback-Leibler (KL) \citep{peters2010relative,schulman2015trust} divergence between new and old
policy and it is well known that the Fisher information matrix, used
to compute the natural gradient is a second order approximation of the
KL divergence. Such trust-region optimization is common in policy
search and has been successfully used to optimize neural network
policies.

However, many properties of the natural gradient are still
under-explored, such as compatible value function approximation \citep{Sutton:1999:PGM:3009657.3009806} for
neural networks, the approximation quality of the KL-divergence and the
online performance of the natural gradient. We analyze the convergence
of the natural gradient analytically and empirically and show that the
natural gradient does not give fast convergence properties if we do
not add an entropy regularization term. This entropy regularization
term results in a new update rule which ensures that the policy 
looses entropy at the correct pace, leading to convergence to a good policy. We further
show that the natural gradient is the optimal (and not the
approximate) solution to a trust region optimization problem for
log-linear models if the natural parameters of the distribution are
optimized and we use compatible value function approximation.

We analyze compatible value function approximation for neural
networks and show that the components of this approximation are
composed of two terms, a state value function which is subtracted from
a state-action value function. While it is well known that the
compatible function approximation denotes an advantage function, the
exact structure was unclear. We show that using compatible value
function approximation, we can derive similar algorithms to Trust
Region Policy Search that obtain the policy update in closed form.
A summary of our contributions is as follows:
\begin{itemize}
\item It is well known that the second-order Taylor approximation to
  trust-region optimization with a KL-divergence bound leads to an
  update direction identical to the natural gradient. However, what is not known is that when using the natural
  parameterization for an exponential policy and using compatible features we can compute the step-size
  for the natural gradient that solves the trust-region update
  exactly for the log-linear parameters.
\item When using an entropy bound in addition to the
  common KL-divergence bound, the compatible features allow us to compute the
  exact update for the trust-region problem in the log-linear case and
  for a Gaussian policy with a state independent covariance we can compute the exact update for the covariance also in the non-linear case.
\item Our new algorithm called Compatible Policy Search (COPOS), based on the above insights, outperforms comparison methods in both continuous control and partially observable discrete action experiments due to entropy control allowing for principled exploration.
\end{itemize}

\section{Preliminaries}

This section discusses background information needed to understand our compatible policy search approach. We first go into Markov decision process (MDP) basics and introduce 
the optimization objective. We continue by showing how trust region methods can help with challenges in updating the policy by using a KL-divergence bound, continue with the classic policy gradient update, introduce the natural gradient and the connection to the KL-divergence bound. Moreover, we introduce the compatible value function approximation and connect it to the natural gradient. Finally, this section concludes by showing how the optimization problem resulting from using an entropy bound to
control exploration can be solved.

Following standard notation, we denote an infinite-horizon discounted Markov decision process (MDP) 
by the tuple 
$(\mathcal{S}, \mathcal{A}, p, r, p_0, \gamma)$,
where $\mathcal{S}$ is a finite set of states and $\mathcal{A}$ is a finite set of actions. $p(s_{t+1} | s_t, a_t)$ denotes
the probability of moving from state $s_t$ to $s_{t+1}$ when the agent executes action $a_t$
at time step $t$. We assume $p(s_{t+1} | s_t, a_t)$ is stationary and unknown but that we can sample from $p(s_{t+1} | s_t, a_t)$ either in simulation or from a physical system. $p_0(s)$ denotes the initial state distribution, $\gamma \in (0,1)$ the discount factor, and $r(s_t,a_t)$ denotes the real valued reward in state $s_t$ when agent executes action $a_t$. The goal is to find a stationary policy $\pi(a_t|s_t)$ that maximizes the expected reward $\mathbb{E}_{s_0, a_0, \dots} \left[ \sum_{t=0}^{\infty} \gamma^t r(s_t, a_t) \right ]$, where
$s_0 \sim p_0(s)$, $s_{t+1} \sim p(s_{t+1} | s_t, a_t)$ and $a_t \sim \pi(a_t| s_t)$.
In the following we will use the notation for the continuous case, where $\mathcal{S}$ and $\mathcal{A}$ denote finite dimensional real valued vector spaces and $\vec s$ denotes the real-valued state vector and
$\vec a$ the real-valued action vector. For discrete states and actions integrals can be replaced in the following by sums.

The expected reward can be defined as~\citep{schulman2015trust}
\begin{equation} 
  J(\pi) = \iint p_{\pi}(\vec{s})\pi(\vec a| \vec s) \Qpi(\vec s, \vec a) d\vec s d\vec a,
  \label{eq:objective}
\end{equation}
where $p_{\pi}(\vec{s})$ denotes a (discounted) state distribution induced by policy $\pi$ and
\begin{flalign*}
  \Qpi(\vec{s}_t, \vec{a}_t) &= \mathbb{E}_{s_{t+1}, a_{t+1}, \dots} \left[ \sum_{t=0}^{\infty} \gamma^t r(s_t, a_t) \right],&\\
  \Vpi(\vec{s}_t) &= \mathbb{E}_{a_t, s_{t+1},\dots} \left[ \sum_{t=0}^{\infty} \gamma^t r(s_t, a_t) \right],
  \Api(\vec{s}_t, \vec{a}_t) =  \Qpi(\vec{s}_t, \vec{a}_t) - \Vpi(\vec{s}_t)&
\end{flalign*}
denote the state-action value function $\Qpi(\vec{s}_t, \vec{a}_t)$, value function $\Vpi(\vec{s}_t)$, and advantage function $\Api(\vec{s}_t, \vec{a}_t)$.

The goal in policy search is to find a policy $\pi(\vec a| \vec s)$ that maximizes Eq.~(\ref{eq:objective}). Usually, policy search computes in each iteration
a new improved policy $\pi$ based on samples generated using the old policy $\pi_{\textrm{old}}$ since
maximizing Eq.~(\ref{eq:objective}) directly is too challenging.
However, since the estimates for
$p_{\pi}(\vec{s})$ and $\Qpi(\vec s, \vec a)$ are based on the old policy, that is, $p_{\piold(\vec s)}$ and $\Qpiold(\vec s, \vec a)$ are actually used in Eq.~(\ref{eq:objective}), they may not be valid for
the new policy. A solution to this is to use Trust-Region Optimization methods which keep the new policy 
sufficiently close to the old policy. Trust-Region Optimization for policy search was first introduced in the relative entropy
policy search (REPS) algorithm \citep{peters2010relative}. Many
variants of this algorithm exist
\citep{akrour2016model-free,Abdolmaleki_NIPS2015,Daniel2016JMLR,akrour2018model}. All
these algorithms use a bound for the KL-divergence of the policy
update which prevents the policy update from being unstable as the new
policy will not go too far away from areas it has not seen
before. Moreover, the bound prevents the policy from being too
greedy. Trust region policy optimization
(TRPO)~\citep{schulman2015trust} uses this bound to optimize
neural network policies.
The policy update can be formulated as finding a policy that maximizes the objective in Eq.~(\ref{eq:objective}) under the KL-constraint:
\begin{align} 
  \textrm{argmax}_{\pi} \mathbb{E}_{p_{\piold}(\vec s)}\big[\int \pi(\vec a| \vec s)  \Qpioldsa d\vec a \big] \\
   \quad \textrm{s.t. } \mathbb{E}_{p_{\piold}(\vec s)}\big[ \textrm{KL}\big(\pi(\cdot| \vec s) || \pi_{\textrm{old}}(\cdot| \vec s) \big) \big] \le \epsilon ,
\label{eq:objective_KL_bound}
\end{align}
where $Q^{\pi_{\textrm{old}}}(\vec s,\vec a)$ denotes the future accumulated reward values of the old policy $\piold$, $p_{\piold}(\vec s)$ the (discounted) state distribution under the old policy, and $\epsilon$ is a constant hyper-parameter. For an $\epsilon$ small enough the state-value and state distribution estimates generated using the old policy are valid also for the new policy since the new and old policy are sufficiently close to each other.

\textbf{Policy Gradient.}
We consider parameterized policies $\pi_{\vec \theta}(\vec a | \vec s)$ with parameter vector $\vec
\theta$. A policy can be improved by modifying the policy parameters in the direction of the policy gradient
which is computed w.r.t.~Eq.~(\ref{eq:objective}).
The "vanilla" policy gradient
\citep{williams1992simple,Sutton:1999:PGM:3009657.3009806} obtained by
the likelihood ratio trick is given by
\begin{align*}
  \nabla_{\vec \theta} J_{\textrm{PG}} &= \iint p(\vec s) \pi_{\vec \theta}(\vec a|
  \vec s) \nabla \log \pi_{\vec \theta}(\vec a| \vec s) \Qpisa d
  \vec s d \vec a \\
  &\approx \sum_i \nabla \log \pi_{\vec
    \theta}(\vec a_i| \vec s_i) \Qpioldsai.
\end{align*} 
The Q-values can be computed by Monte-Carlo estimates (high variance),
that is, $\Qpioldsat \approx \sum_{h = 0}^\infty \gamma^h r_{t + h}$ or
estimated by policy evaluation techniques (typically high bias). We
can further subtract a state-dependent baseline $V(\vec s)$ which
decreases the variance of the gradient estimate while leaving it
unbiased, that is,
\begin{align*}
  \nabla_{\vec \theta} J_{\textrm{PG}} \approx \sum_i \nabla \log \pi_{\vec \theta}(\vec a_i| \vec s_i) \big(\Qpioldsai - V(\vec s_i)\big).
\end{align*}

\textbf{Natural Gradient.}
Contrary to the "vanilla" policy gradient, the natural gradient~\citep{amari98natural} method uses the steepest descent direction in a Riemannian manifold, so it is effective in learning, avoiding plateaus.
The natural gradient can be obtained by using a
second order Taylor approximation for the KL divergence, that is,
$\mathbb{E}_{p(\vec s)}\left[\textrm{KL}\big(\pi_{\vec \theta + \vec
    \alpha}(\cdot|\vec s)|| \pi_{\vec \theta}(\cdot|\vec s)\big)
  \right] \approx \vec \alpha^T \vec F \vec \alpha,$
where $\vec F$ is the Fisher information matrix~\citep{amari98natural}.
The natural gradient is now defined as the update direction that is
most correlated with the standard policy gradient and has a limited
approximate KL, that is,
\begin{equation*}
  \nabla_{\vec \theta} J_{\textrm{NAC}} = \textrm{argmax}_{\vec \alpha}
  \vec \alpha^T \nabla_{\vec \theta} J_{\textrm{PG}} \quad\textrm{s.t.}\quad \vec
  \alpha^T \vec F \vec \alpha \le \epsilon
\end{equation*}
resulting in 
\begin{equation*}
  \nabla_{\vec \theta} J_{\textrm{NAC}} = \eta^{-1} \vec
  F^{-1} \nabla_{\vec \theta} J_{\textrm{PG}},
\end{equation*}
where $\eta$ is a Lagrange multiplier.

\textbf{Compatible Value Function Approximation.}
It is well known that we can obtain an unbiased gradient with
typically smaller variance if compatible value function approximation
is used \citep{Sutton:1999:PGM:3009657.3009806}. An approximation of
the Monte-Carlo estimates $\tilde{G}^{\pi_\textrm{old}}_{\vec w}(\vec
s, \vec a) = \vec \phi(\vec s, \vec a)^T \vec w$ is compatible to the
policy $\pi_{\vec \theta}(\vec a| \vec s)$, if the features $\vec
\phi(\vec s, \vec a)$ are given by the log gradients of the policy,
that is, $\vec \phi(\vec s, \vec a) = \nabla_{\vec \theta} \log \pi_{\vec
  \theta}(\vec a| \vec s)$. The parameter $\vec w$ of the
approximation $\tilde{G}^{\pi_\textrm{old}}_{\vec w}(\vec s, \vec a)$
is the solution of the least squares problem
\begin{equation*}
  \vec w^* = \textrm{argmin}_{\vec w} \; \mathbb{E}_{p(\vec
  s)\pi_{\vec \theta}(\vec a|\vec s)}\left[
  \left(Q^{\pi_{\textrm{old}}}(\vec s, \vec a) - 
  \vec \phi(\vec s, \vec a)^T \vec w \right)^2 \right].
\end{equation*}
\citet{4863} showed that in the case of compatible value function
approximation, the inverse of the Fisher information matrix cancels
with the matrix spanned by the compatible features and, hence,
$\nabla_{\vec \theta} J_{\textrm{NAC}} = \eta^{-1} \vec
w^*$. 
Another interesting observation is
that the compatible value function approximation is in fact not an
approximation for the Q-function but for the advantage function
$A^{\pi_\textrm{old}}(\vec s, \vec a) = Q^{\pi_\textrm{old}}(\vec s,
\vec a) - V^{\pi_\textrm{old}}(\vec s)$ as the compatible features are
always zero mean. In
Section~\ref{sec:compatible_natural_policy_search}, we show how 
with compatible value function approximation the natural gradient
directly gives us an exact solution to the trust region optimization 
problem instead of requiring a search for the update step size
to satisfy the KL-divergence bound in trust region optimization.

\textbf{Entropy Regularization.}
Recently, some approaches \citep{Abdolmaleki_NIPS2015,akrour2016model-free,mnih2016asynchronous,DBLP:journals/corr/ODonoghueMKM16} use an additional bound for the entropy of the resulting policy. The entropy bound can be beneficial since it
allows to limit the change in exploration potentially preventing greedy policy convergence.
The trust
region problem is in this case given by
\begin{align} 
  &\textrm{argmax}_{\pi} \mathbb{E}_{p_{\textrm{old}}(\vec s)}\big[\int \pi(\vec a|
    \vec s) \Qpioldsa d\vec a \big] \nonumber &\\ 
  \quad \textrm{s.t.}\quad&
  \mathbb{E}_{p_{\textrm{old}}(\vec s)}\big[ \textrm{KL}\big(\pi(\cdot| \vec s) ||
    \pi_{\textrm{old}}(\cdot| \vec s) \big) \big] \le \epsilon \nonumber &\\
  &\mathbb{E}_{p_{\textrm{old}}(\vec
    s)}\left[H(\pi_{\textrm{old}}(\cdot|\vec s)) - H(\pi(\cdot|\vec
    s))\right] \le \beta, & \label{eq:objective_with_entropy}
\end{align}
where the second constraint limits the expected loss in entropy ($H()$ denotes Shannon entropy in the discrete case and differential entropy in the continuous case) for
the new distribution (applying an entropy constraint only on $\pi(\vec
a|\vec s)$ but adjusting $\beta$ according to $\pi_{\textrm{old}}(\vec
a|\vec s)$ is equivalent in \citep{akrour2016model-free,akrour2018model}). 
The policy update rule can be formed for the constrained optimization problem
by using the method of Lagrange multipliers:
\begin{equation} 
  \pi(\vec a|\vec s) \propto \pi_{\textrm{old}}(\vec a|\vec
  s)^{{\frac{\eta}{\eta + \omega}}}\exp\left( \frac{\Qpioldsa}{\eta + \omega} \right),
  \label{eq:more_policy_update}
\end{equation}
where $\eta$ and $\omega$ are Lagrange multipliers
\citep{akrour2016model-free}. $\eta$ is associated with the KL-divergence
bound $\epsilon$ and $\omega$ is related to the entropy bound $\beta$.
Note that, for $\omega = 0$, the entropy
bound is not active and therefore, the solution is equivalent to the
standard trust region solution. It has been realized that the entropy
bound is needed to prevent premature convergence issues connected with
the natural gradient. We show that these premature convergence issues
are inherent to the natural gradient as it always reduces entropy of
the distribution. In contrast, entropy control can prevent this.

\section{Compatible Policy Search with Natural Parameters}
\label{sec:compatible_natural_policy_search}
In this section, we analyze the natural gradient update equations for
exponential family distributions. This analysis reveals an important
connection between the natural gradient and the trust region
optimization: Both are equivalent if we use the natural
parameterization of the distribution in combination with compatible
value function approximation. This is an important insight as the natural gradient now provides the optimal solution for a given trust region, not just an approximation which is commonly believed: for example, \citet{schulman2015trust} have to use line search to fit the
natural gradient update to the KL-divergence bound. Moreover, this insight
can be applied together with the entropy bound to control policy exploration
and get a closed form update in the case of compatible log-linear policies.
Furthermore, the use of compatible value function
approximation has several advantages in terms of variance reduction
which can not be achieved with the plain Monte-Carlo estimates which
we leave for future work. We also present an analysis of the online
performance of the natural gradient and show that entropy
regularization can converge exponentially faster. Finally, we present
our new algorithm for Compatible Policy Search (\Acronym{}) which uses
the insights above.

\subsection{Equivalence of Natural Gradients and Trust Region Optimization}
We first consider soft-max distributions that are log-linear in
the parameters (for example, Gaussian distributions or the Boltzmann
distribution) and subsequently extend our results to non-linear
soft-max distributions, for example given by neural networks. A
log-linear soft-max distribution can be represented as
\begin{equation*}
  \pi(\vec a|\vec s) = \frac{\exp\left(\vec \psi(\vec s, \vec a)^T
    \vec \theta\right)}{\int \exp\left(\vec \psi(\vec s, \vec a)^T \vec
    \theta\right) d \vec a}.
\end{equation*} 
Note that also Gaussian distributions can be represented this way (see, for example,
Eq.~(\ref{eq:gaussian_natural})), however, the natural parameterization is
commonly not used for Gaussian distributions. Typically, the Gaussian
is parameterized by the mean $\vec \mu$ and the covariance matrix $\vec
\Sigma$. However, the natural parameterization and our analysis suggest
that the precision matrix $\vec B = \vec \Sigma^{-1}$ and the linear
vector $\vec b = \vec \Sigma^{-1} \vec \mu$ should be used to benefit
from many beneficial properties of the natural gradient.

It makes sense to study the exact form of the compatible approximation
for these log-linear models. The compatible features are given by
\begin{equation*}
  \vec \phi(\vec s, \vec a) = \nabla_{\vec \theta} \log \pi_{\vec
    \theta}(\vec a| \vec s) = \vec \psi(\vec s, \vec a) -
  \mathbb{E}_{\pi(\cdot|\vec s)}\left[\vec \psi(\vec s,
    \cdot)\right]. 
\end{equation*}
As we can see, the compatible feature space is
always zero mean, which is inline with the observation that the
compatible approximation $\tilde{G}^{\pi_\textrm{old}}_{\vec w}(\vec
s, \vec a)$ is an advantage function. Moreover, the structure of the
features suggests that the advantage function is composed of a term
for the Q-function $\tilde{Q}_{\vec w}(\vec s, \vec a) = \vec
\psi(\vec s, \vec a)^ T\vec w$ and for the value function
$\tilde{V}_{\vec w}(\vec s) = \mathbb{E}_{\pi(\cdot|\vec
  s)}\left[\tilde{Q}_{\vec w}(\vec s, \vec a)\right]$, that is,
\begin{align*}
  \tilde{G}^{\pi_\textrm{old}}_{\vec w}(\vec s, \vec a) =
  \vec \psi(\vec s, \vec a)^ T\vec w
 - \mathbb{E}_{\pi(\cdot|\vec s)}\left[\vec
    \psi(\vec s, \cdot)^T\vec w\right]
\end{align*}
We can now directly use the compatible advantage function
$\tilde{G}^{\pi_\textrm{old}}_{\vec w}(\vec s, \vec a)$ in our trust
region optimization problem given in Eq.~(\ref{eq:objective_KL_bound}).
The resulting policy is then given by
\begin{align*}
  \pi(\vec a| \vec s) & \propto \pi_{\textrm{old}}(\vec a| \vec s)
  \exp\left(\frac{\vec \psi(\vec s, \vec a)^T\vec w - 
    \mathbb{E}_{\pi(\cdot|\vec s)}\left[\vec \psi(\vec s, \cdot)^T\vec w\right]}
           {\eta} \right)\\
  & \propto \exp\big(\psi(\vec s, \vec a)^ T (\vec \theta_{\textrm{old}} +
    \eta^{-1} \vec w)\big).
\end{align*}
Note that the value function part of $\tilde{G}_{\vec w}$ does not
influence the updated policy. Hence, if we use the natural
parameterization of the distributions in combination with compatible
function approximation, then we directly get a parametric update of
the form
\begin{equation}
  \vec \theta = \vec \theta_{\textrm{old}}  + \eta^{-1} \vec w .
  \label{eq:policy_update_KL}
\end{equation}
Furthermore, the suggested update is equivalent to the natural
gradient update: The natural gradient is the {\em optimal solution}
for a given trust region problem and not just an
approximation. However, this statement only holds if we use natural
parameters and compatible value function approximation. Moreover, the
update needs only the Q-function part of the compatible function
approximation.

We can do a similar analysis for the optimization problem with entropy
regularization given in Eq.~(\ref{eq:objective_with_entropy}) using Eq.~(\ref{eq:more_policy_update}). The
optimal policy given the compatible value function approximation is
now given by
\begin{equation}
  \pi(\vec a| \vec s) \propto \exp\left(\psi(\vec s, \vec a)^ T 
           \left(\frac{\eta \vec \theta_{\textrm{old}} + \vec w}
                      {\eta + \omega}\right)\right), \quad \text{yielding} \quad
  \vec \theta = \frac{\eta \vec \theta_{\textrm{old}}  + \vec w}{\eta + \omega}.
  \label{eq:policy_update_loglinear}
\end{equation}
In comparison to the standard natural gradient, the influence of the
old parameter vector is diminished by the factor $\eta / (\eta +
\omega)$ which will play an important role for our further analysis.

\subsection{Compatible Approximation for Neural Networks}

So far, we have only considered models that are log-linear in the
parameters (ignoring the normalization constant). For more complex
models, we need to introduce non-linear parameters $\vec \beta$ for
the feature vector, that is, $\vec \psi(\vec s, \vec a) = \vec \psi_{\vec
  \beta}(\vec s, \vec a)$. We are in particular interested in Gaussian
policies in the continuous case and softmax policies in the discrete
case as they are the standard for continuous and discrete actions,
respectively. In the continuous case, we could either use a Gaussian
with a constant variance where the mean is parameterized by a neural
network, a non-linear interpolation linear feedback controllers with
Gaussian noise or also Gaussians with state-dependent variance. For
simplicity, we will focus on Gaussians with a constant covariance
$\vec \Sigma$ where the mean is a product of neural network (or any
other non-linear function) features $\varphi_i(\vec s)$ and a mixing
matrix $\vec K$ that could be part of the neural network output
layer. The policy and the log policy are then
\begin{flalign}
  \pi(\vec a| \vec s) &= \mathcal{N}\left(\vec a \Bigg| \sum_i \varphi_i(\vec s) \vec k_i, \vec \Sigma \right) & \label{eq:gaussian_policy}\\
  \log \pi(\vec a| \vec s) &=
  - 0.5 \vec\varphi(\vec s)^T \vec K^T \vec\Sigma^{-1} \vec K \vec\varphi(\vec s) 
  + \vec \varphi(\vec s)^T \vec K^T \vec \Sigma^{-1} \vec a
  - 0.5 \vec a^T \vec \Sigma^{-1} \vec a + \textrm{const} & \nonumber \\
  & = 
  - 0.5 \vec\varphi(\vec s)^T \vec U \vec K \vec\varphi(\vec s) +
  \vec \varphi(\vec s)^T \vec U \vec a  
  - 0.5 \vec a^T \vec \Sigma^{-1} \vec a + \textrm{const}, & \label{eq:gaussian_natural}
\end{flalign}
where $\vec K = (\vec k_1, \dots, \vec k_N)$ and $\vec U = \vec K^T
\vec\Sigma^{-1}$. To compute $\vec \psi(\vec s, \vec a)$ we note that
\begin{flalign*}
  \vec \phi(\vec s, \vec a) &= \nabla_{\vec \theta} \log \pi_{\vec
    \theta}(\vec a| \vec s) = \vec \psi(\vec s, \vec a) -
  \mathbb{E}_{\pi(\cdot|\vec s)}\left[\vec \psi(\vec s, \cdot)\right]
\end{flalign*}
To get $\vec \psi(\vec s, \vec a)$ we note that some parts of Eq.(\ref{eq:gaussian_natural}) and thus of $\nabla_{\vec \theta} \log \pi_{\vec
    \theta}(\vec a| \vec s)$ do not depend on $\vec a$.
We ignore those parts for computing $\vec \psi(\vec s, \vec a)$ since 
$\vec \psi(\vec s, \vec a) \vec \theta$ is the state-action value function and
action independent parts of the state-action value function do not influence the optimal action choice. Thus we get
\begin{equation}
  \vec \psi(\vec s, \vec a) = \nabla_{\vec \theta} \left(\vec \varphi(\vec s)^T \vec U \vec a - 0.5 \vec a^T \vec \Sigma^{-1} \vec a\right).
\end{equation}

We then take the gradient w.r.t.\ the log-linear parameters $\vec
\theta = (\vec\Sigma^{-1}, \vec U)$ resulting in
$\nabla_{\vec\Sigma^{-1}} \log \hat{\pi}(\vec a| \vec s) = - 0.5 \vec
a \vec a^T$, $\nabla_{\vec U} \log \hat{\pi}(\vec a| \vec s) = \vec a
\vec \varphi(\vec s)^T$, and $\vec \psi(\vec s, \vec a) = [-
  \vecto[0.5 \vec a \vec a^T], \vecto[\vec a \vec \varphi(\vec
    s)^T]]^T$, where $\vecto[\cdot]$ concatenates matrix columns into
a column vector.

Note that the variances and the linear parameters of the mean are
contained in the parameter vector $\vec \theta$ and can be updated by
the update rule in Eq.~(\ref{eq:policy_update_loglinear}) explained above. However, for obtaining the update rules for
the non-linear parameters $\vec \beta$, we first have to compute the
compatible basis, that is,
\begin{equation}
  \nabla_{\vec \beta} \log \pi_{\vec \beta,\vec \theta}(\vec a|\vec s)
  = \partial \vec \psi_{\vec \beta}(\vec s, \vec a) \vec \theta / \partial \vec
  \beta - \mathbb{E}_{\pi(\cdot|\vec
    s)}[\partial \vec \psi_{\vec \beta}(\vec s, \cdot) \vec \theta / \partial \vec
    \beta].
    \label{eq:nonlinear_basis}
\end{equation}
Note that due to the $\log$ operator the derivative is linear w.r.t.\ log-linear
parameters $\vec \theta$. For the Gaussian distribution in Eq.~(\ref{eq:gaussian_policy}) the
gradient of the action dependent parts of the log policy in
Eq.~(\ref{eq:nonlinear_basis}) become
\begin{align*}
  \partial \vec \psi_{\vec \beta}(\vec s, \vec a) \vec \theta / \partial \vec
    \beta &=
    \frac{\partial}{\partial \vec \beta}
  \vec \varphi(\vec s)^T \vec U \vec a.
\end{align*}
Now, in order to find the update rule for the non-linear parameters we will write
the update rule for the policy using Eq.~(\ref{eq:more_policy_update}), and, using 
the value function formed by multiplying the compatible basis in Eq.~(\ref{eq:nonlinear_basis})
by $\vec w_{\beta}$ which is the part of the compatible
approximation vector that is responsible for $\vec \beta$:
\begin{align}
  \pi_{\vec \beta,\vec \theta}(\vec a|\vec s) \propto &
  \; \pi_{\vec \beta_{\textrm{old}},\vec \theta_{\textrm{old}}}(\vec a|\vec s)^{\eta / (\eta + \omega)} 
  \exp \Big(\left(\nabla_{\vec \beta_{\textrm{old}}} \vec \psi_{\vec \beta_{\textrm{old}}}(\vec s, \vec a) \vec \theta_{\textrm{old}}\right) \vec w_{\vec \beta} - &\nonumber\\
  & \; \mathbb{E}_{\pi(\cdot|\vec
      s)}\left[\left(\nabla_{\vec \beta_{\textrm{old}}} \vec \psi_{\vec \beta_{\textrm{old}}}(\vec s, \vec a) \vec \theta_{\textrm{old}}\right)\right] \vec w_{\vec \beta}\Big) &\label{eq:1}\\
  \propto &
  \; \exp\left(\frac{\eta \vec \psi_{\vec \beta_{\textrm{old}}}(\vec s, \vec a) \vec \theta_{\textrm{old}}}
  					{\eta + \omega}\right)
     \exp\left(\frac{
     \nabla_{\vec \beta_{\textrm{old}}} (\vec \psi_{\vec \beta_{\textrm{old}}}(\vec s, \vec a) \vec \theta_{\textrm{old}})
     \vec w_{\vec \beta}}{\eta + \omega} \right) &\label{eq:2}\\
  = &
  \;\exp\left(\frac{\eta \vec \psi_{\vec \beta_{\textrm{old}}}(\vec s, \vec a) \vec \theta_{\textrm{old}} +
    \nabla_{\vec \beta_{\textrm{old}}} (\vec \psi_{\vec \beta_{\textrm{old}}}(\vec s, \vec a) \vec \theta_{\textrm{old}})
    \vec w_{\vec \beta}}{\eta + \omega} \right) &\nonumber\\
  = & 
  \;\exp\left(\frac{\eta}{\eta + \omega} \left(
    \vec \psi_{\vec \beta_{\textrm{old}}}(\vec s, \vec a) +
    \nabla_{\vec \beta_{\textrm{old}}} (\vec \psi_{\vec \beta_{\textrm{old}}}(\vec s, \vec a)\vec w_{\vec \beta} / \eta)\right)
    \vec \theta_{\textrm{old}} \right) &\label{eq:3}\\
  \approx &
  \;\exp\left(\frac{\eta}{\eta + \omega}
    \vec \psi_{\vec \beta_{\textrm{old}} + \vec w_{\beta} / \eta}(\vec s, \vec a)
    \vec \theta_{\textrm{old}} \right), &\label{eq:4}
\end{align}
where we dropped action independent parts, which can be seen as part of the distribution normalization, from Eq.~(\ref{eq:1}) to Eq.~(\ref{eq:2}). Note that
Eq.~(\ref{eq:4}) represents the first order Taylor
approximation of Eq.(~\ref{eq:3}) at $\vec w_{\vec \beta} / \eta = 0$.
Moreover, note that rescaling
of the energy function $\vec \psi_{\vec \beta}(\vec s, \vec a) \vec \theta$ is implemented by the update of the parameters
$\vec \theta$ and hence can often be ignored for the update for $\vec \beta$.
The approximate update rule for $\vec \beta$ is thus
\begin{equation}
  \vec \beta = \vec \beta_{\textrm{old}} + \vec w_{\vec \beta} / \eta .
  \label{eq:policy_update_nonlinear}
\end{equation}
Hence, we can conclude that the natural gradient is an approximate
trust region solution for the non-linear parameters $\vec \beta$ as
the first order Taylor approximation of the energy function is
replaced by the real energy function after the update. Still, for the
parameters $\vec \theta$, which in the end dominate the mean and
covariance of the policy, the natural gradient is the exact trust
region solution.

\begin{algorithm}[t]
  \caption{Compatible Policy Search (\Acronym{}).}
  \label{alg:copos}
  \begin{algorithmic}
    \State Initialize $\pi_0$
  \For{i = $1$ \textbf{to} $\textrm{max episodes}$}
    \State Sample $(\vect{s},\vect{a},r)$ tuples using $\pi_{i-1}$
    \State Estimate advantage function $A_{\pi_{i-1}}(s,a)$ from samples
    \State Solve $\vect{w} = \vect{F}^{-1} \nabla J_{PG}(\pi_{i})$
    \State Use compatible value function to solve Lagrange multipliers $\eta$ and $\omega$ for Eq.~(\ref{eq:objective_with_entropy})
    \State Update $\pi_i$ using $\vect{w}$, $\eta$ and $\omega$ based on Eq.~(\ref{eq:policy_update_loglinear}) and Eq.~(\ref{eq:policy_update_nonlinear})
  \EndFor
  \end{algorithmic}
\end{algorithm}

\subsection{Compatible Value Function Approximation in Practice}
\label{sec:CompatibleValueFunctionApproximationInPractice}
Algorithm~\ref{alg:copos} shows the Compatible Policy Search (COPOS)
approach (see Appendix~\ref{sec:technical_details_for_experiments} for a more detailed description of the
discrete action algorithm version). In COPOS, for the policy updates
in Eq.~(\ref{eq:policy_update_loglinear})
and Eq.~(\ref{eq:policy_update_nonlinear}), we need to find $\vec w$,
$\eta$, and $\omega$. For estimating $\vec w$ we could use the
compatible function approximation.
In this paper, we do not estimate the value function explicitly but
instead estimate $\vec w$ as a natural gradient using the conjugate
gradient method which removes the need for computing the inverse of
the Fisher information matrix explicitly (see
for example~\citep{schulman2015trust}). As discussed before $\eta$ and
$\omega$ are Lagrange multipliers associated with the KL-divergence
and entropy bounds. In the log-linear case with compatible natural
parameters, we can compute them exactly using the dual of the
optimization objective Eq.~(\ref{eq:objective_with_entropy}) and in the
non-linear case approximately. In the continuous action case, the
basic dual includes integration over both actions and states but we
can integrate over actions in closed form due to the compatible value
function: we can eliminate terms which do not depend on the
action. The dual resolves into an integral over just states allowing
computing $\eta$ and $\omega$ efficiently. Please, see Appendix~\ref{sec:lagrange_solution} for more details. Since $\eta$ is an approximation for the
non-linear parameters, we performed in the experiments for the
continuous action case an additional alternating optimization twice:
1) we did a binary search to satisfy the KL-divergence bound while
keeping $\omega$ fixed, 2) we re-optimized $\omega$ (exactly, since
$\omega$ depends only on log-linear parameters) keeping $\eta$
fixed. For discrete actions it was sufficient to perform only an
additional line search to update the non-linear parameters.

\section{Analysis and Illustration of the Update Rules}
\label{sec:analysis_update_rules}

We will now analyze the performance of both update rules, with and
without entropy, in more detail with a simple stateless Gaussian
policy $\pi(a) \propto \exp(-0.5 B a^2 + b a)$ for a scalar action
$a$. Our real reward function $R(a) = -0.5 R a^2 + r a$ is also
quadratic in the actions. We assume an infinite number of samples to
perfectly estimate the compatible function approximation. In this case
$\tilde{G}_{\vec w}$ is given by $\tilde{G}_{\vec w}(a) = R(a) = -0.5
R a^2 + r a$ and $\vec w = [R,r]$.
The reward function maximum is $a^* = R^{-1}r$.

\textbf{Natural Gradients.}

For now, we will analyze the performance of the natural gradient if
$\eta$ is a constant and not optimized for a given KL-bound. After $n$
update steps, the parameters of the policy are given by $B_n = B_0 + n R / \eta$, $b_n = b_0 + n r / \eta$.
The distance between the mean $\mu_n = B_n^{-1}
b_n$ of the policy and the optimal solution $a^*$ is
\begin{equation*}
  d_n = \mu_n - a^* = \frac{b_0 R - r B_0}{R (B_0 + n R / \eta)} =
  O\left(\frac{c_0}{c_1 + c_2 n} \right).
\end{equation*}
We can see that the learned solution approaches the optimal
solution, however, very slowly and heavily
depending on the precision $B_0$ of the initial solution. The reason for
this effect can be explained by the update rule of the precision. As
we can see, the precision $B_n$ is increased at every update
step. This shrinking variance in turn decreases the step-size for the
next update.

\textbf{Entropy Regularization.} Here we provide the derivation of $d_n$ for the entropy regularization
case. We perform a similar analysis for the entropy regularization
update rule. We start with constant parameters $\eta$ and $\omega$ and
later consider the case with the trust region. The distance $d_n = \mu_n - a^*$ is again a function of $n$. The updates for the
entropy regularization result in the following parameters after $n$
iterations
\begin{equation*}
  B_n = B_0 \frac{\eta^n}{(\eta + \omega)^n} + R / \eta \left(1 -
  \frac{\eta^n}{(\eta + \omega)^n} \right) / \left(1 - \frac{\eta}{\eta
    + \omega} \right),
\end{equation*}
\begin{equation*}
  b_n = b_0 \frac{\eta^n}{(\eta + \omega)^n} + r / \eta \left(1 -
  \frac{\eta^n}{(\eta + \omega)^n} \right) / \left(1 - \frac{\eta}{\eta
    + \omega} \right).
\end{equation*}
The distance $d_n = \mu_n - a^*$ can again be expressed as a function of $n$:
\begin{equation}
  d_n =
  \frac{b_0 - r B_0 / R}{B_0 + R ((\eta + \omega)^n
    / \eta^n - 1) (\eta + \omega) / (\eta \omega) } =
  O\left(\frac{c_0}{c_1 + c_2^n} \right),
  \label{eq:distance_actions}
\end{equation}
with $c_2 > 1$. Hence, also this update rule converges to the correct
solution but contrary to the natural gradient, the part of the
denominator that depends on $n$ grows exponentially. As the old
parameter vector is always multiplied by a factor smaller than one,
the influence of the initial precision matrix $B_0$ vanishes while
$B_0$ dominates natural gradient convergence. While the
natural gradient always decreases variance, entropy
regularization avoids the entropy loss and can even increase variance.

\begin{figure}
  \centering
  \tabcolsep=0.0cm
  \begin{tabular}{cccc}	
    \includegraphics[width=0.25\textwidth]{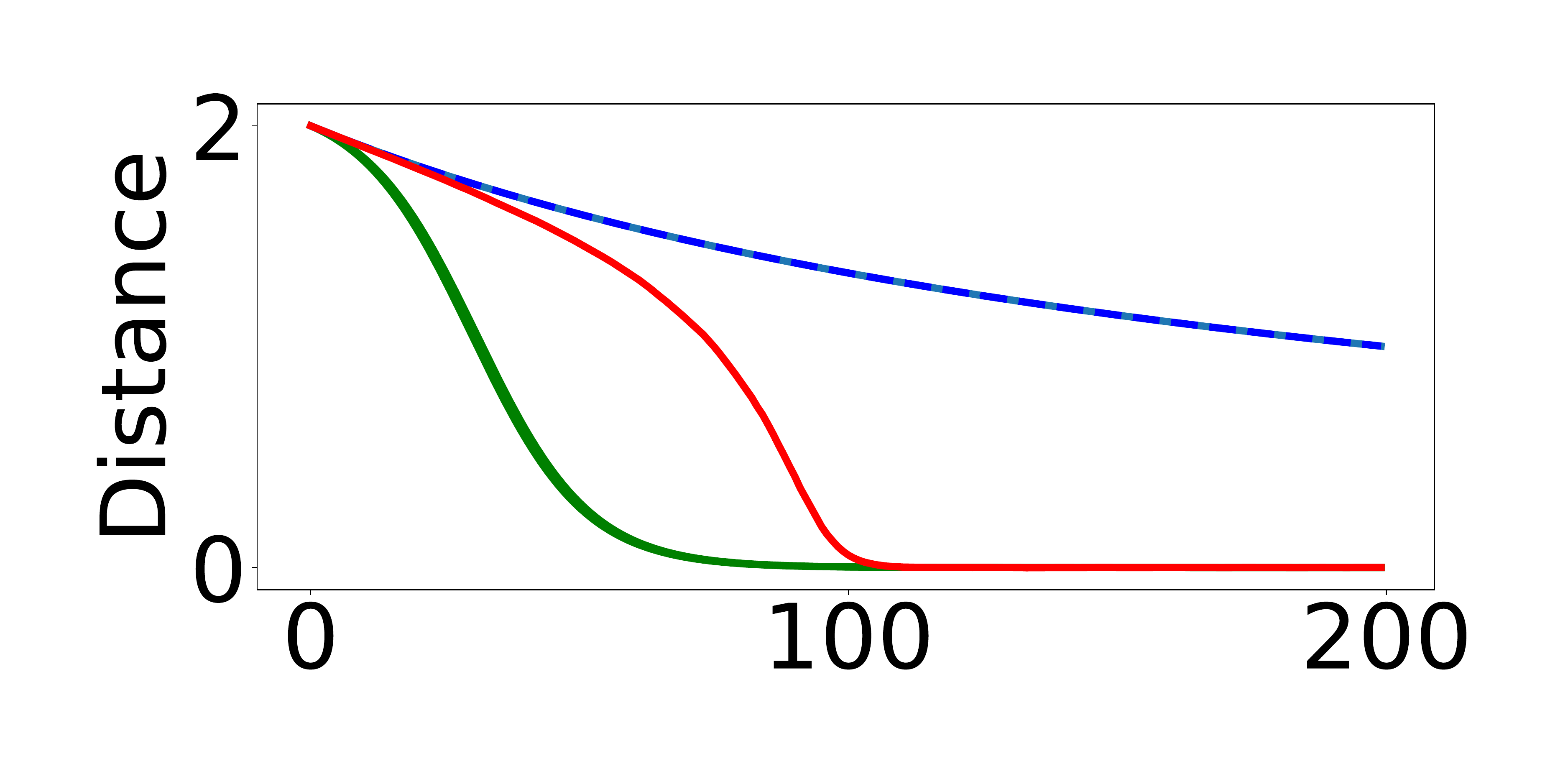} &
    \includegraphics[width=0.25\textwidth]{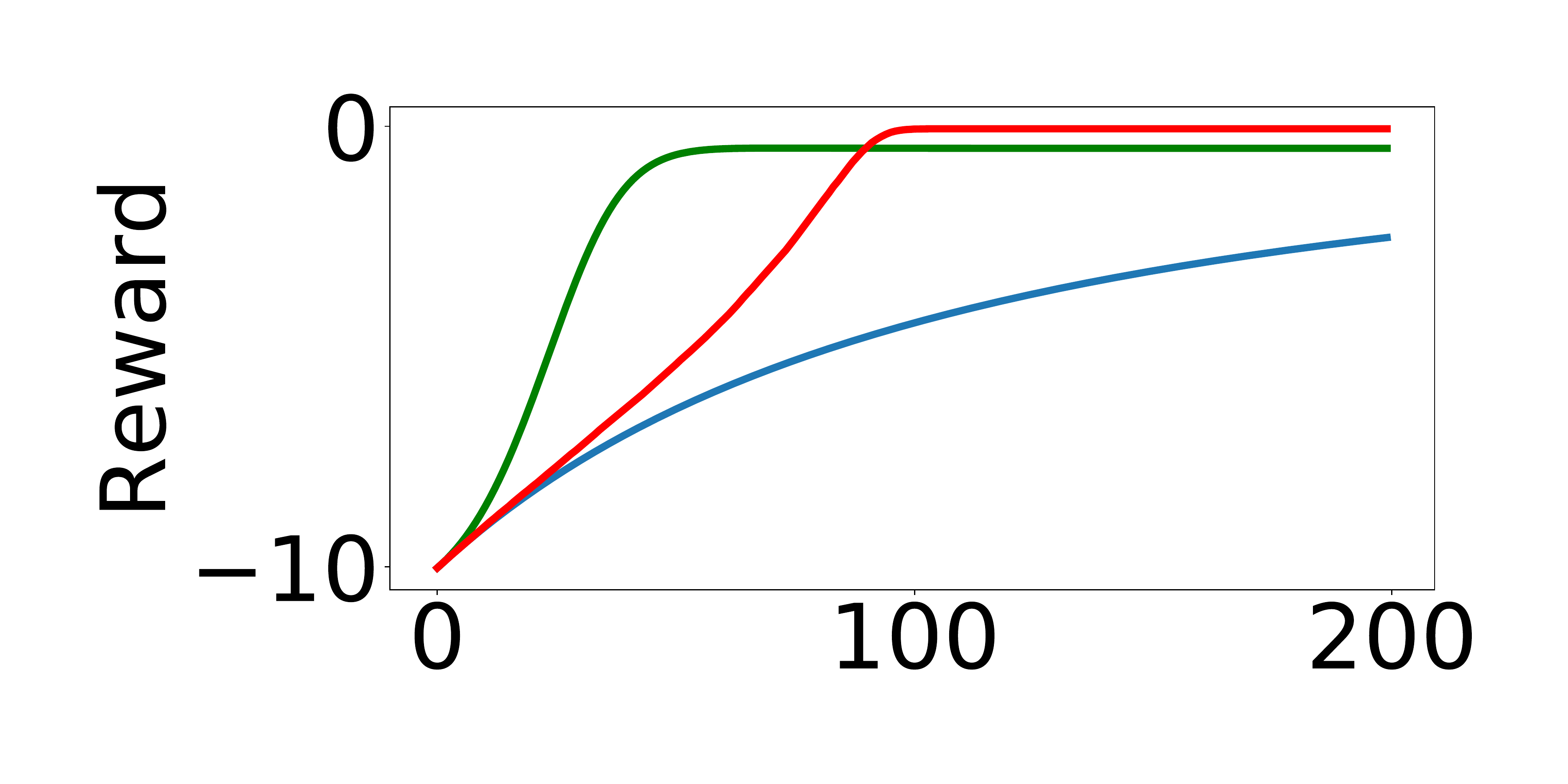} &
    \includegraphics[width=0.25\textwidth]{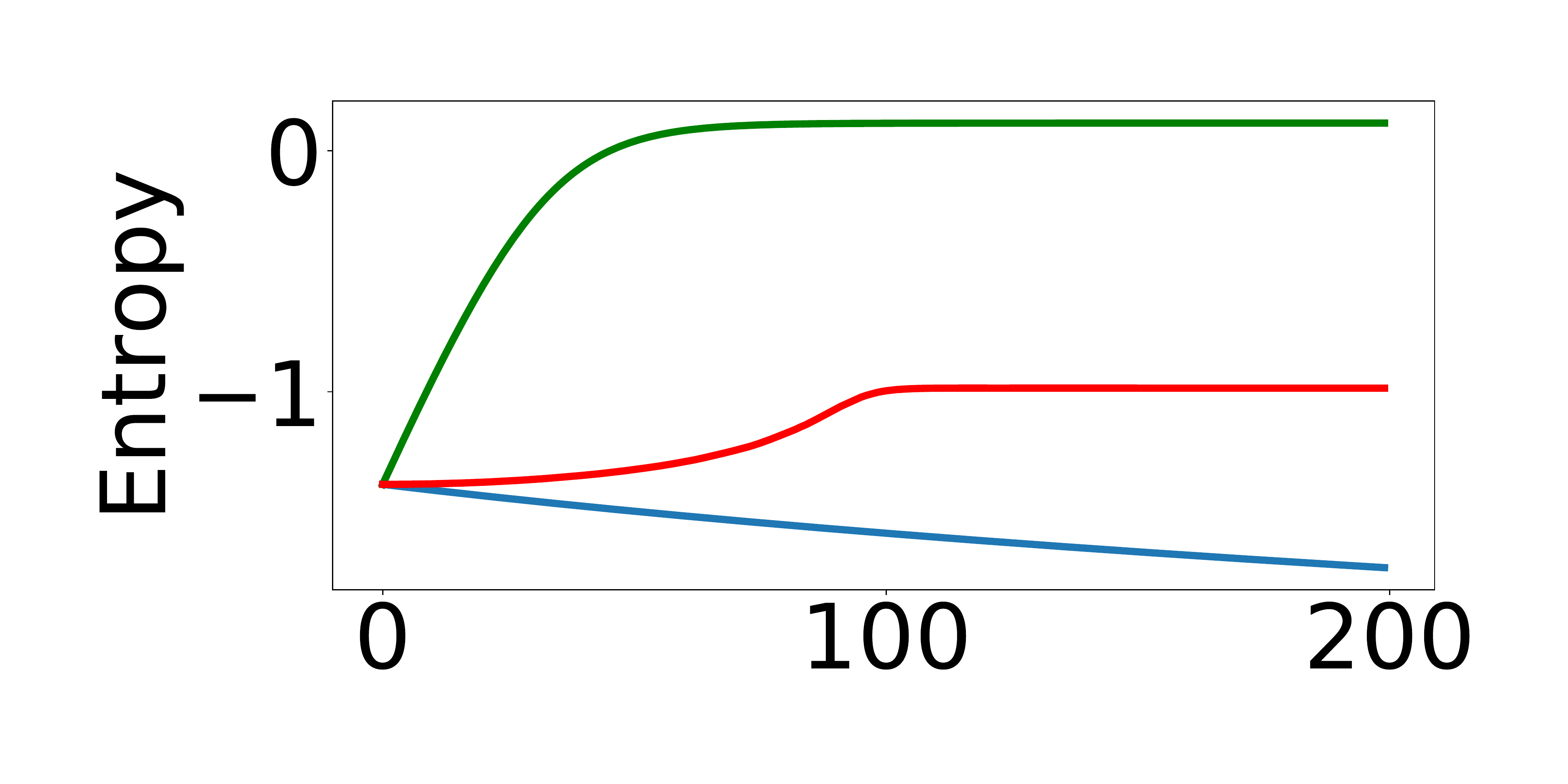} &
    \includegraphics[width=0.25\textwidth]{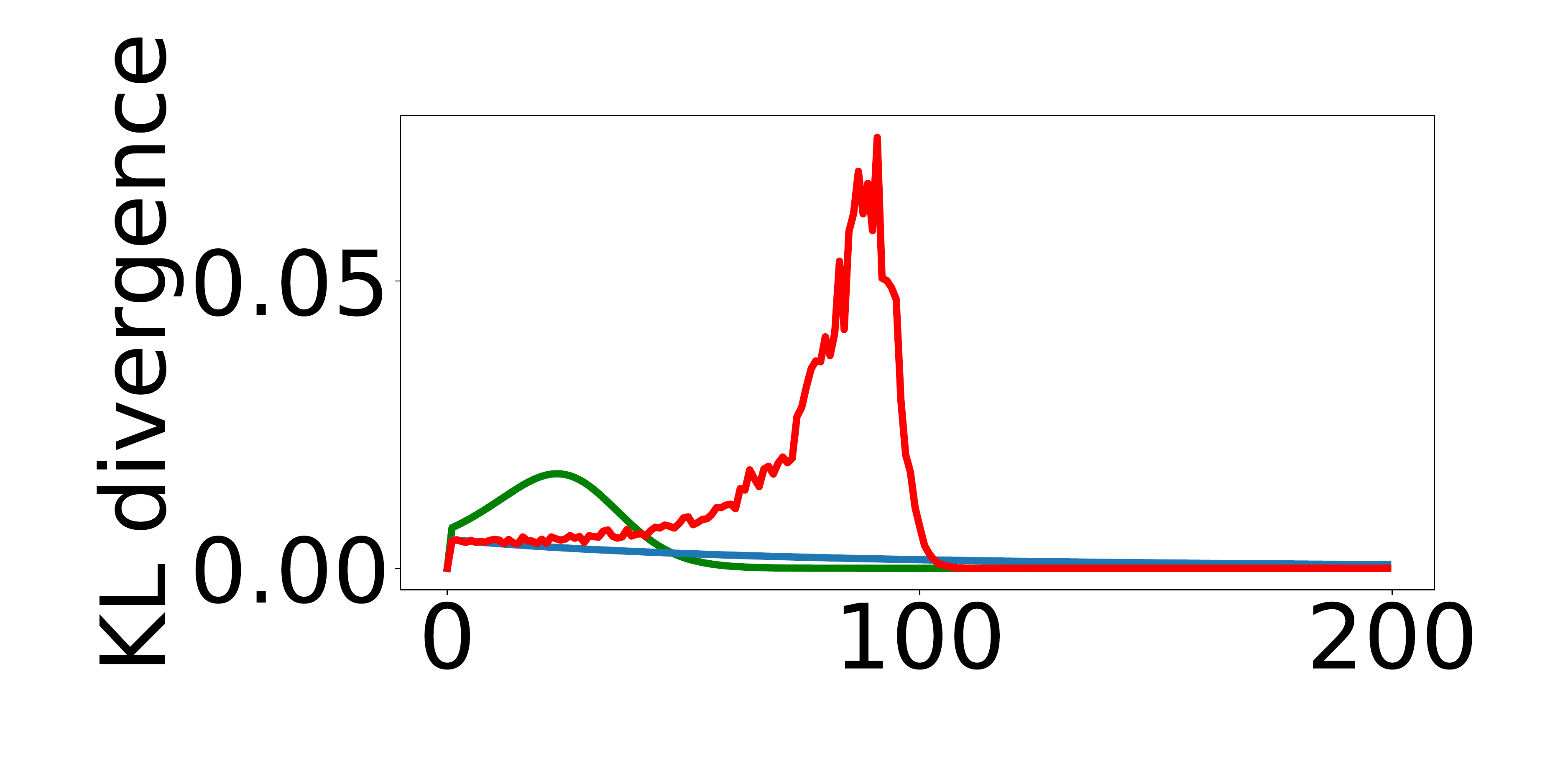} \\
    \includegraphics[width=0.25\textwidth]{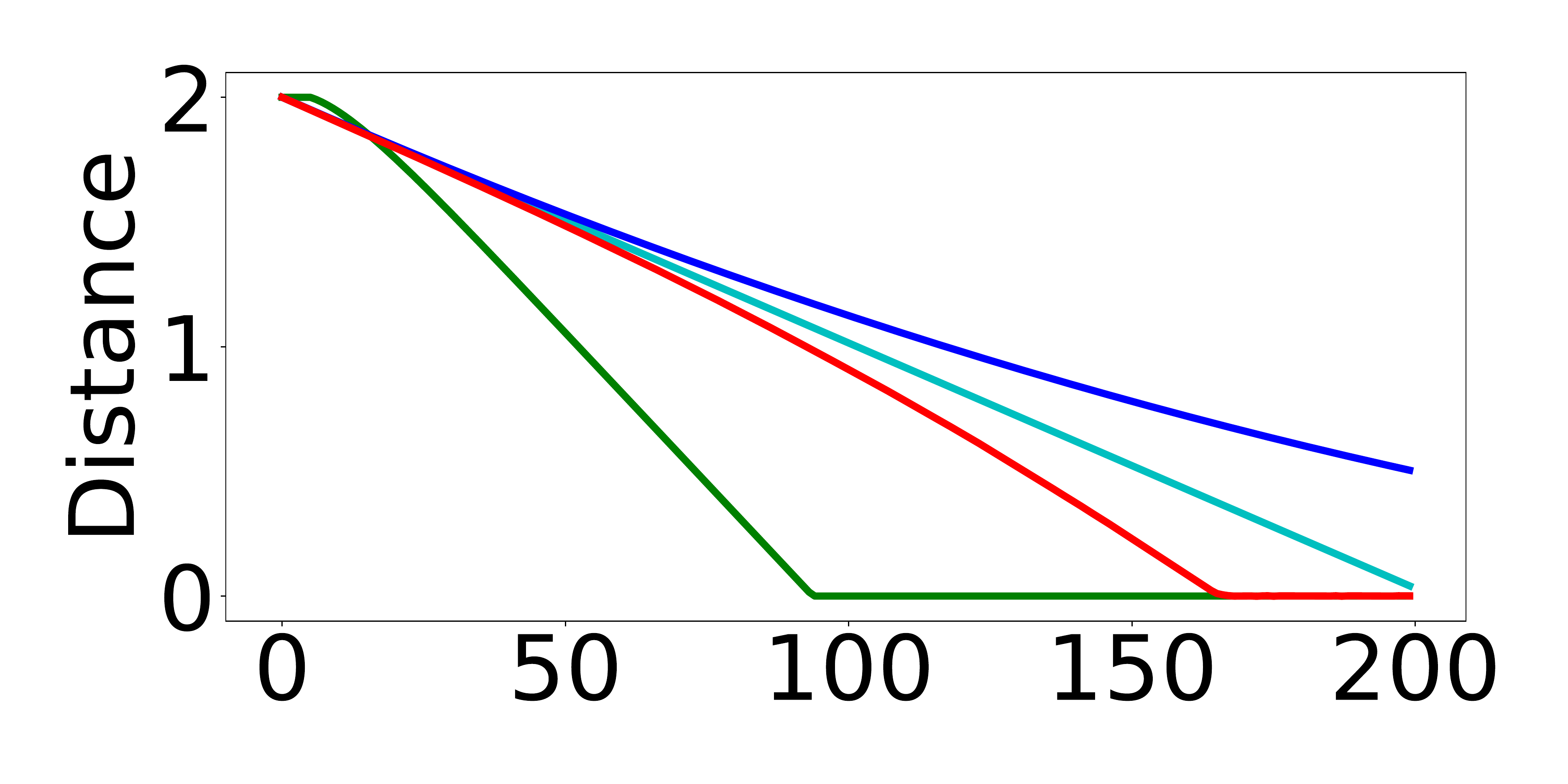} &
    \includegraphics[width=0.25\textwidth]{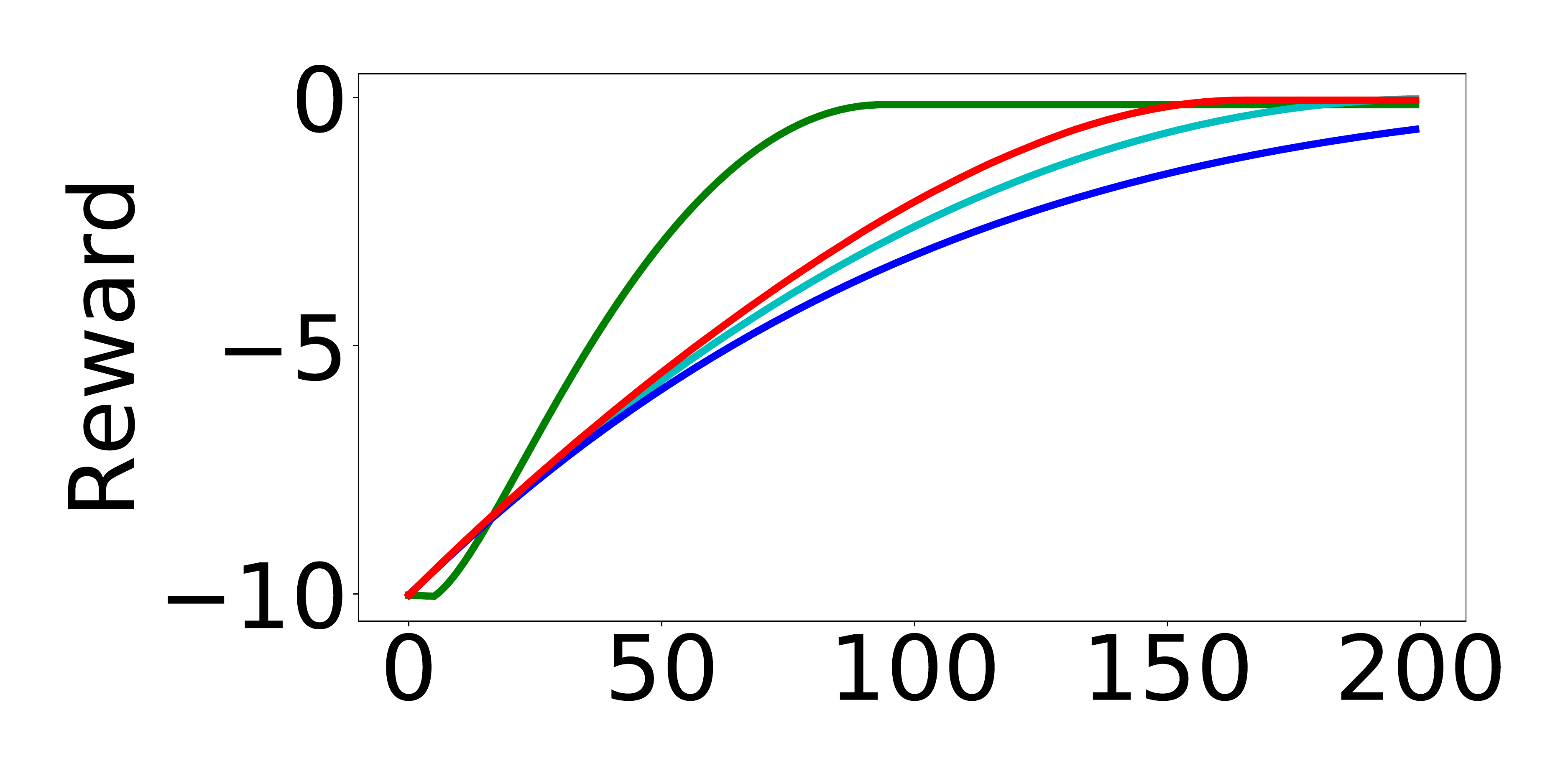} &
    \includegraphics[width=0.25\textwidth]{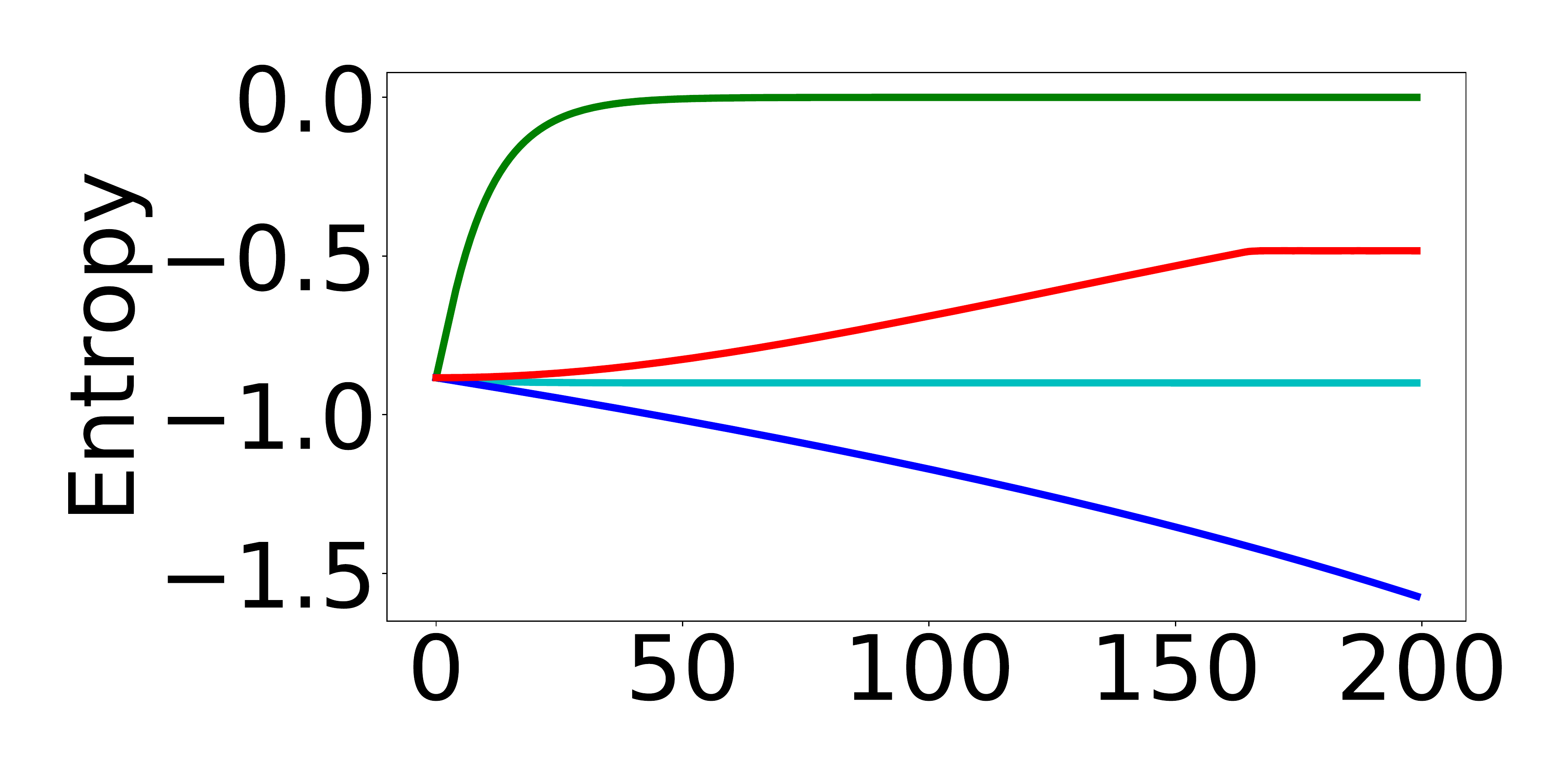} &
    \includegraphics[width=0.25\textwidth]{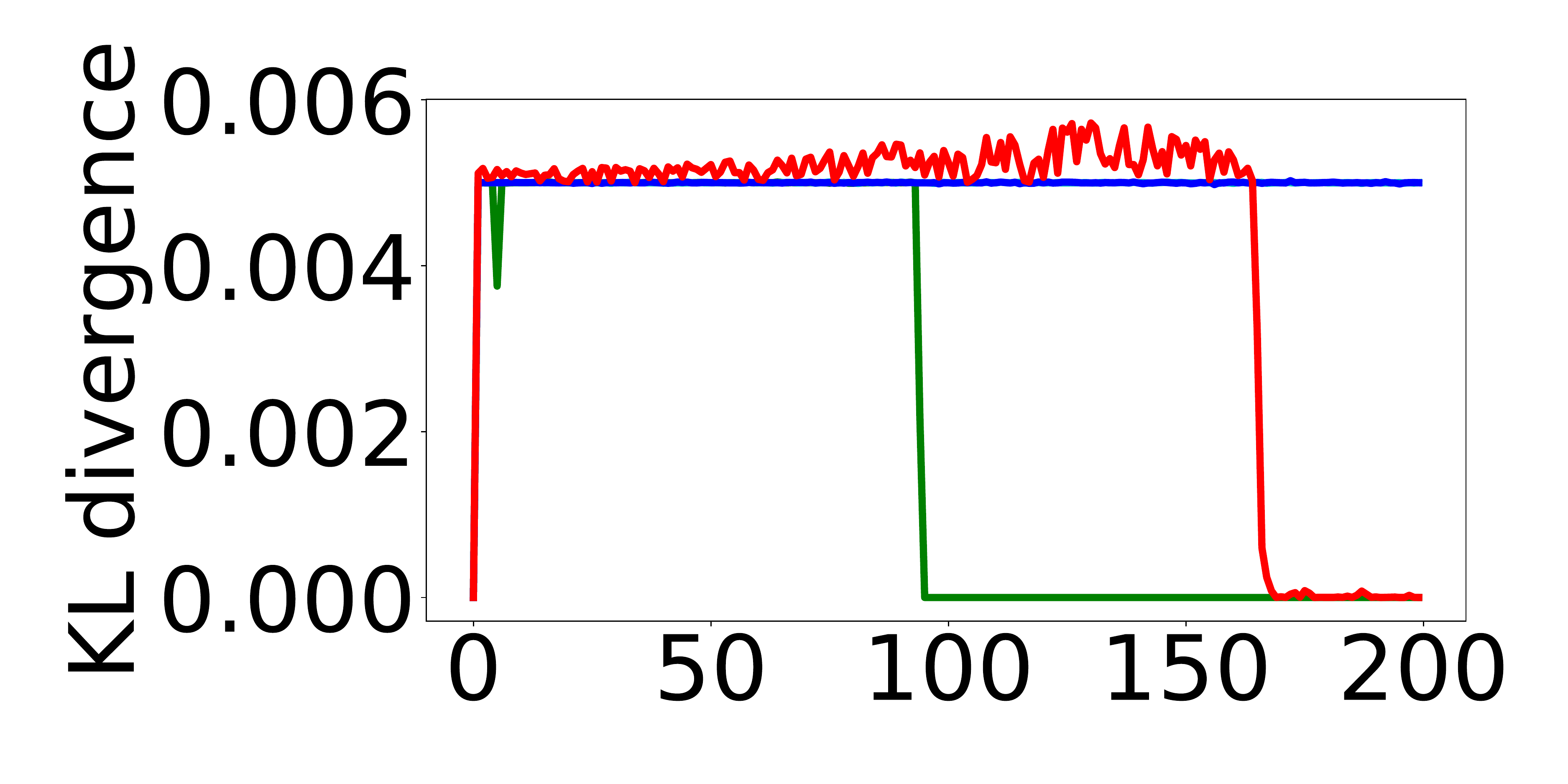}
  \end{tabular}
  \caption{\label{fig:quadratic_updates} Comparison of different
    update rules with and without entropy regularization in the toy
    example of Section~\ref{sec:analysis_update_rules}. The figures
    show the ``Distance'' between optimal and current policy mean (see
    Eq.~(\ref{eq:distance_actions})), the expected ``Reward'', the
    expected ``Entropy'', and the expected ``KL-divergence'' between
    previous and current policy over $200$ iterations
    (x-axis). \textbf{Top:} Policy updates with constant learning
    rates and no trust region. Comparison of the natural gradient
    (blue), natural gradient with entropy regularization (green) and
    vanilla policy gradient (red). \textbf{Bottom:} Policy updates
    with trust region. Comparison of the natural gradient (blue),
    natural gradient where the entropy is controlled to a set-value
    (green), natural gradient with zero entropy loss (cyan) and
    vanilla policy gradient (red). To summarize, without entropy
    regularization the natural gradient decreases the entropy too
    fast.}
\end{figure}

\textbf{Empirical Evaluation of Constant Updates.}  We plotted the
behavior of the algorithms, and standard policy gradient, for this
simple toy task in Figure~\ref{fig:quadratic_updates} (top). We use
$\eta = 10$ and $\omega = 1$ for the natural gradient and the entropy
regularization and a learning rate of $\alpha = 1000$ for the policy
gradient. We estimate the standard policy gradients from 1000 samples.
Entropy regularization performs favorably speeding up learning in the
beginning by increasing the entropy. With constant parameters $\eta$
and $\omega$, the algorithm drives the entropy to a given
target-value. The policy gradient performs better than the natural
gradient as it does not reduce the variance all the time and even
increases the variance. However, the KL-divergence of the standard
policy gradient appears uncontrolled.

\textbf{Empirical Evaluation of Trust Region Updates.} 
In the trust region case, we minimized the Lagrange dual at each
iteration yielding $\eta$ and $\omega$. We chose at each iteration the
highest policy gradient learning rate where the KL-bound was still
met. For entropy regularization we tested two setups: 1) We fixed the
entropy of the policy (that is, $\gamma = 0$), 2) The entropy of the
policy was slowly driven to $0$. Figure~\ref{fig:quadratic_updates}
(bottom) shows the results. The natural gradient still suffers from
slow convergence due to decreasing the entropy of the policy
gradually. The standard gradient again performs better as it increases
the entropy outperforming even the zero entropy loss natural
gradient. For entropy control, even more sophisticated scheduling
could be used such as the step-size control of
CMA-ES~\citep{hansen2001completely} as a heuristic that works well.

\section{Related Work}
\label{sec:related_work}

Similar to classical reinforcement learning the leading contenders in
deep reinforcement learning can be divided into value based-function
methods such as Q-learning with deep Q-Network (DQN)
\citep{mnih2015humanlevel}, actor-critic methods
\citep{wu2017scalable,tangkaratt18,abdolmaleki18}, policy gradient methods such as
deep deterministic policy gradient (DDPG)
\citep{silver2014deterministic,DBLP:journals/corr/LillicrapHPHETS15}
and policy search methods based on information theoretic / trust
region methods, such as proximal policy optimization (PPO)
\citep{DBLP:journals/corr/SchulmanWDRK17} and trust region policy
optimization (TRPO) \citep{schulman2015trust}.

Trust region optimization was introduced in the relative entropy
policy search (REPS) method~\citep{peters2010relative}. TRPO and
TNPG~\citep{schulman2015trust} are the first methods to apply trust
region optimization successfully to neural networks. In contrast
to TRPO and TNPG, we derive our method from the compatible value
function approximation perspective. TRPO and TNPG differ from our
approach, in that they do not use an entropy constraint and do not
consider the difference between the log-linear and non-linear
parameters for their update. On the technical level, compared to TRPO, 
we can update the log-linear parameters (output layer of neural network and the covariance) with an exact update step while TRPO does a line search to 
find the update step. Moreover, for the covariance we can find an exact
update to enforce a specific entropy and thus control exploration while TRPO does not
bound the entropy, only the KL-divergence.
PPO also applies an adaptive KL penalty term.

\citet{Kakade:2001,bagnell2003covariant,4863,geist10} have also
suggested similar update rules based on the natural gradient for the
policy gradient framework. 
\citet{wu2017scalable} applied approximate natural gradient updates to both the actor and critic
in an actor-critic framework but did not utilize compatible value functions or
an entropy bound.
\citet{4863,geist10} investigated the idea
of compatible value functions in combination with the natural gradient
but used manual learning rates instead of trust region
optimization. The approaches in
\citep{Abdolmaleki_NIPS2015,akrour2016model-free} use an entropy bound
similar to ours. However, the approach in \citep{Abdolmaleki_NIPS2015}
is a stochastic search method, that is, it ignores sequential decisions
and views the problem as black-box optimization, and the approach in
\citep{akrour2016model-free} is restricted to trajectory
optimization. Moreover, both of these approaches do not explicitly
handle non-linear parameters such as those found in neural networks.
The entropy bound used in \citep{tangkaratt18} is similar to ours,
however, their method depends on second order approximations of a deep
Q-function, resulting in a much more complex policy update that can
suffer from the instabilities of learning a non-linear Q-function.

For exploration one can in general add an entropy term to the
objective. In the experiments, we compare against TRPO with this additive 
entropy term. In preliminary
experiments, to control entropy in TRPO, we also combined the entropy
and KL-divergence constraints into a single constraint without success.

\section{Experiments}
\label{sec:experiments}
In the experiments, we focused on investigating the following research
question: Does the proposed entropy regularization approach help to
improve performance compared to other methods which do not control the
entropy explicitly? For selecting comparison methods we followed
\citep{DuanCHSA16} and took four gradient based methods: Trust Region
Policy Optimization (TRPO) \citep{schulman2015trust}, Truncated Natural
Policy Gradient \citep{DuanCHSA16,schulman2015trust}, REINFORCE (VPG)
\citep{williams1992simple}, Reward-Weighted Regression (RWR)
\citep{NIPS2008_3545} and two gradient-free black box optimization
methods: Cross Entropy Method (CEM) \citep{Rubinstein1999}, Covariance
Matrix Adaption Evolution Strategy (CMA-ES)
\citep{hansen2001completely}. We used rllab~\footnote{\url{http://rllab.readthedocs.io/en/latest/}} for algorithm implementation.
We ran experiments in both challenging
continuous control tasks and discrete partially observable tasks which
we discuss next.

\begin{figure}[t]
  \centering
  \tabcolsep=0.0cm
  \begin{tabular}{cc}	
    \includegraphics[width=0.5\textwidth]{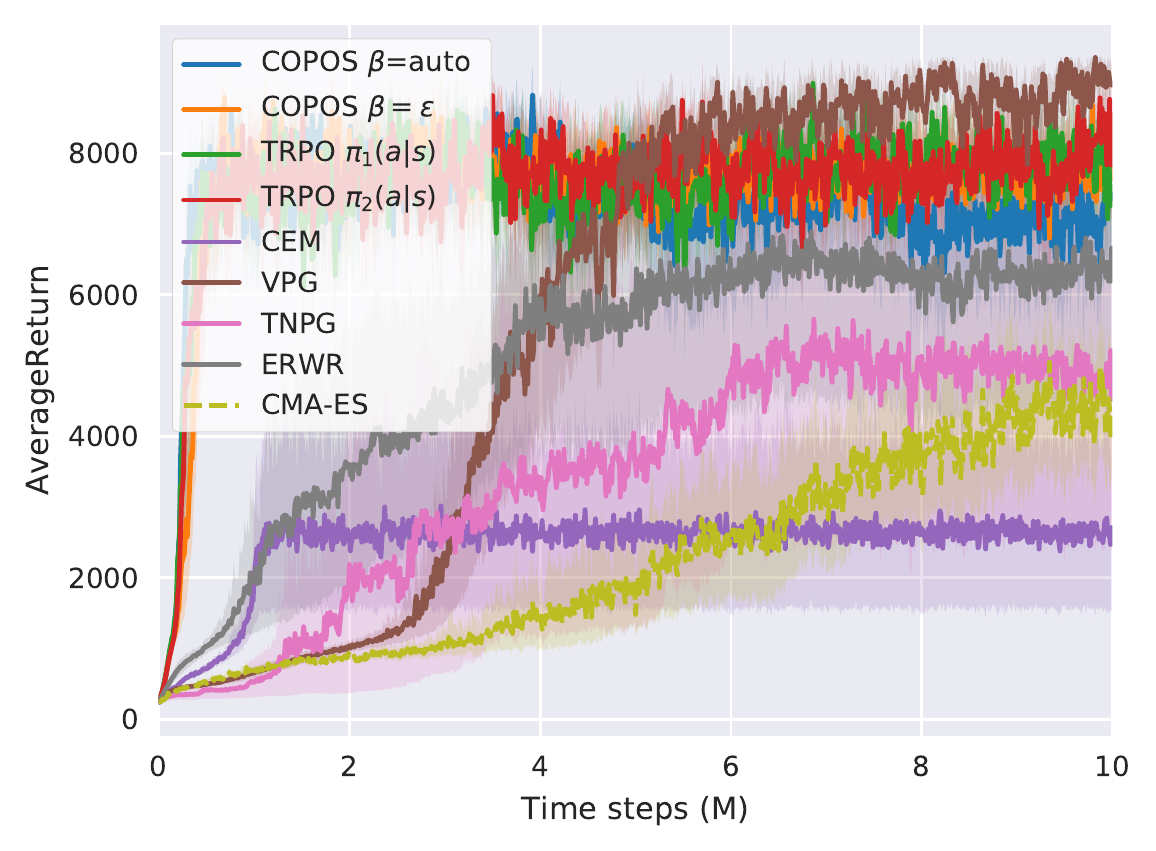}\vspace{-0.3em} &
    \includegraphics[width=0.5\textwidth]{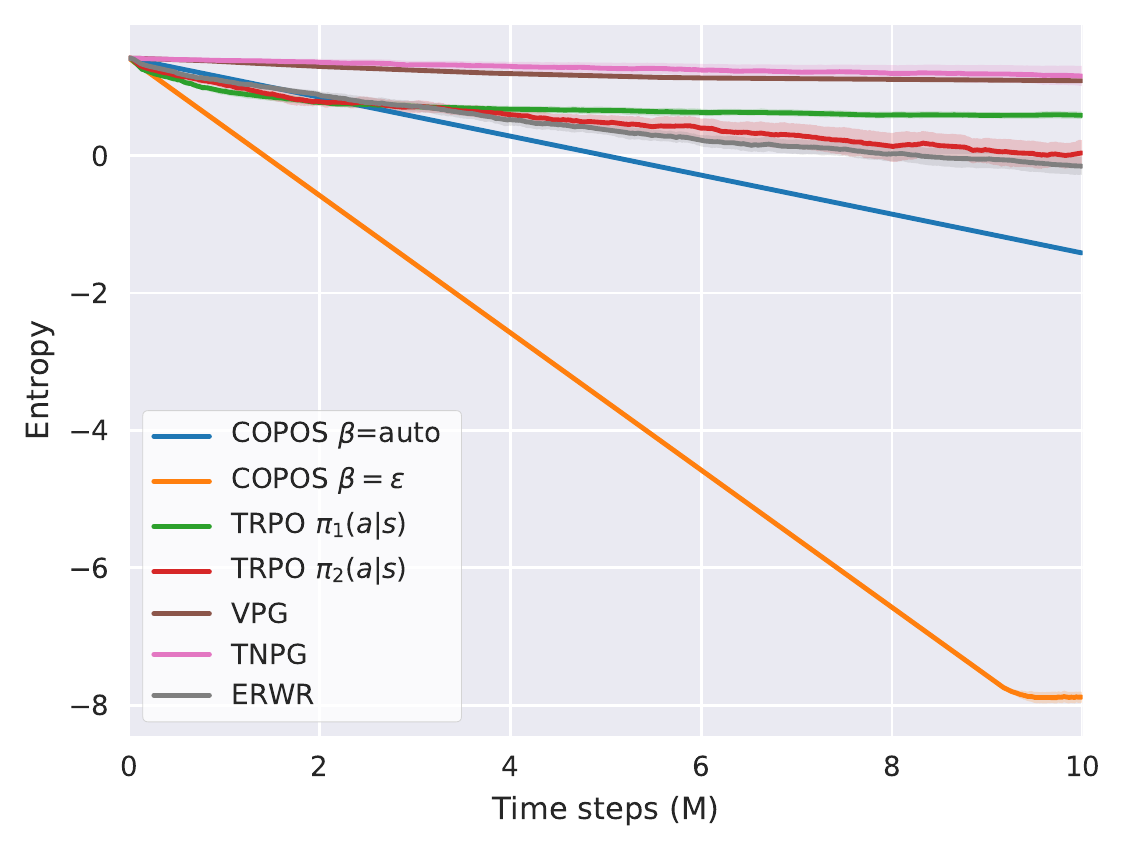}\vspace{-0.3em} \\
    \multicolumn{2}{c}{\scriptsize (a) RoboschoolInvertedDoublePendulum-v1} \\
    \includegraphics[width=0.5\textwidth]{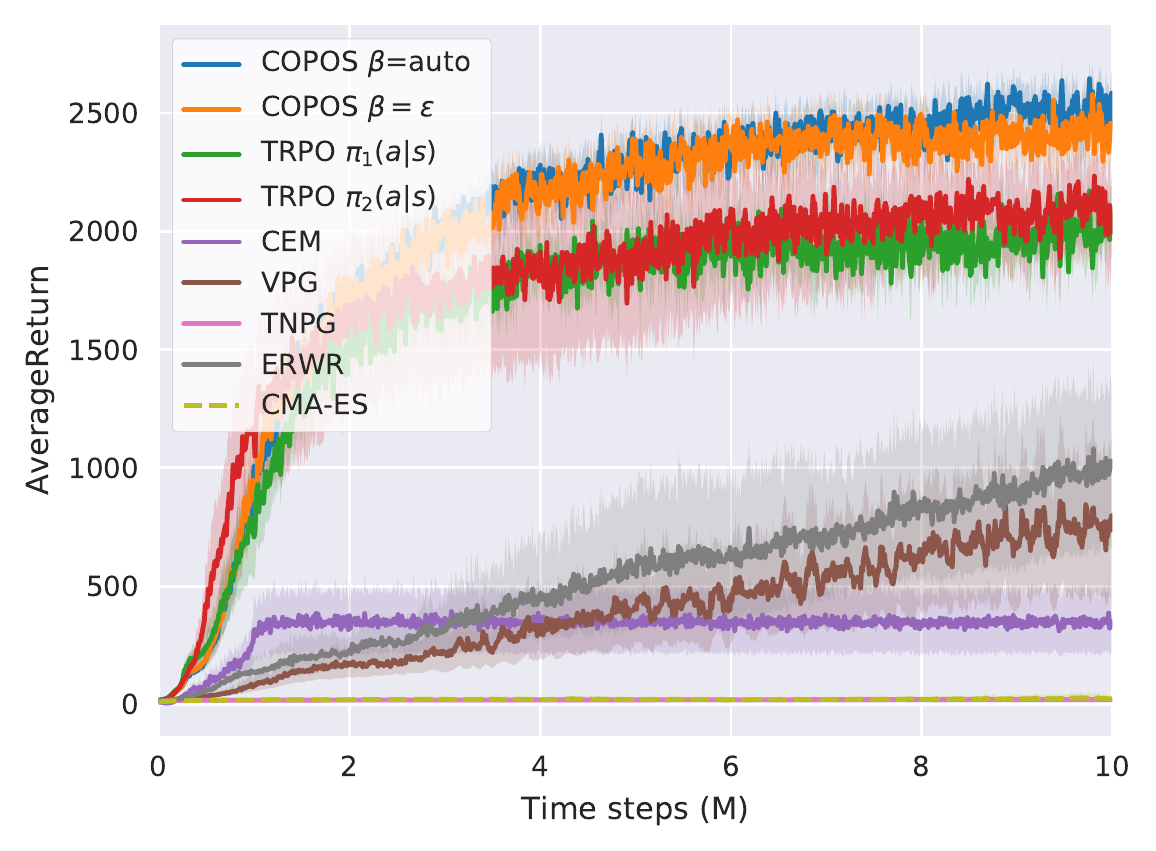}\vspace{-0.3em} &
    \includegraphics[width=0.5\textwidth]{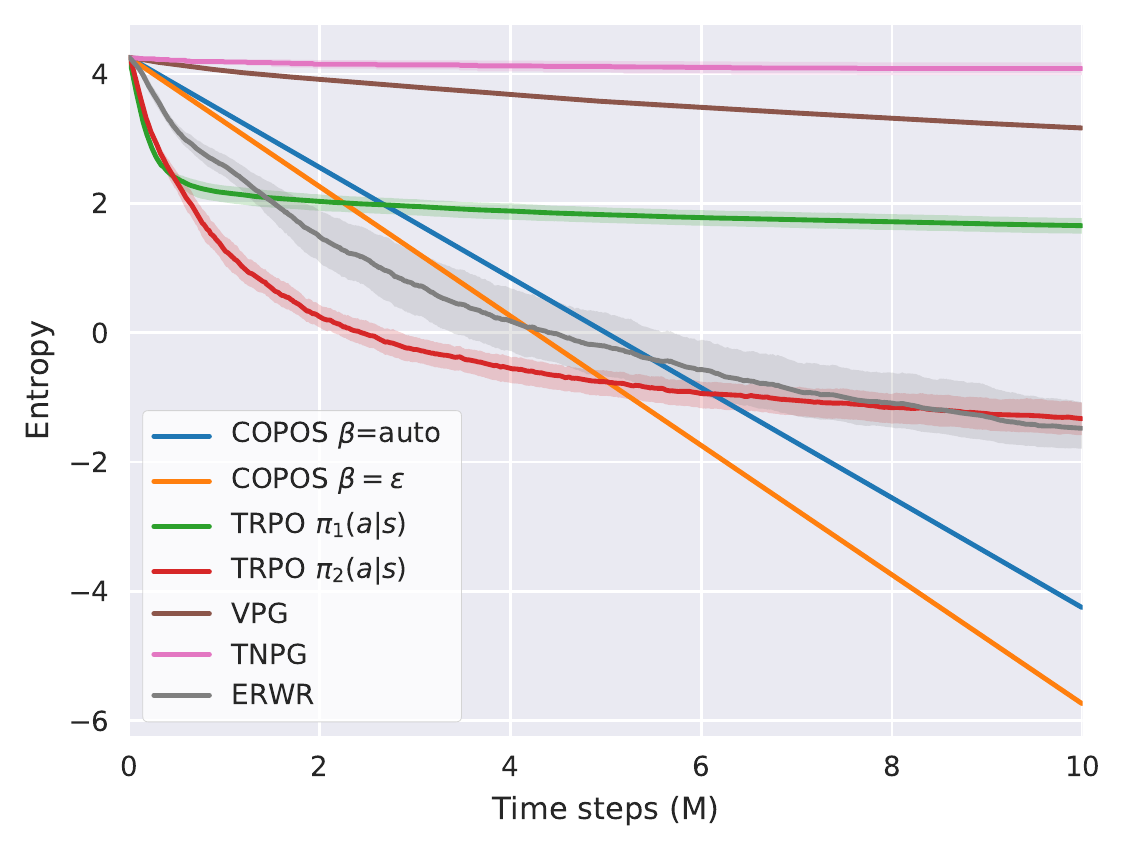}\vspace{-0.3em} \\
    \multicolumn{2}{c}{\scriptsize (b) RoboschoolHopper-v1} \\
    \includegraphics[width=0.5\textwidth]{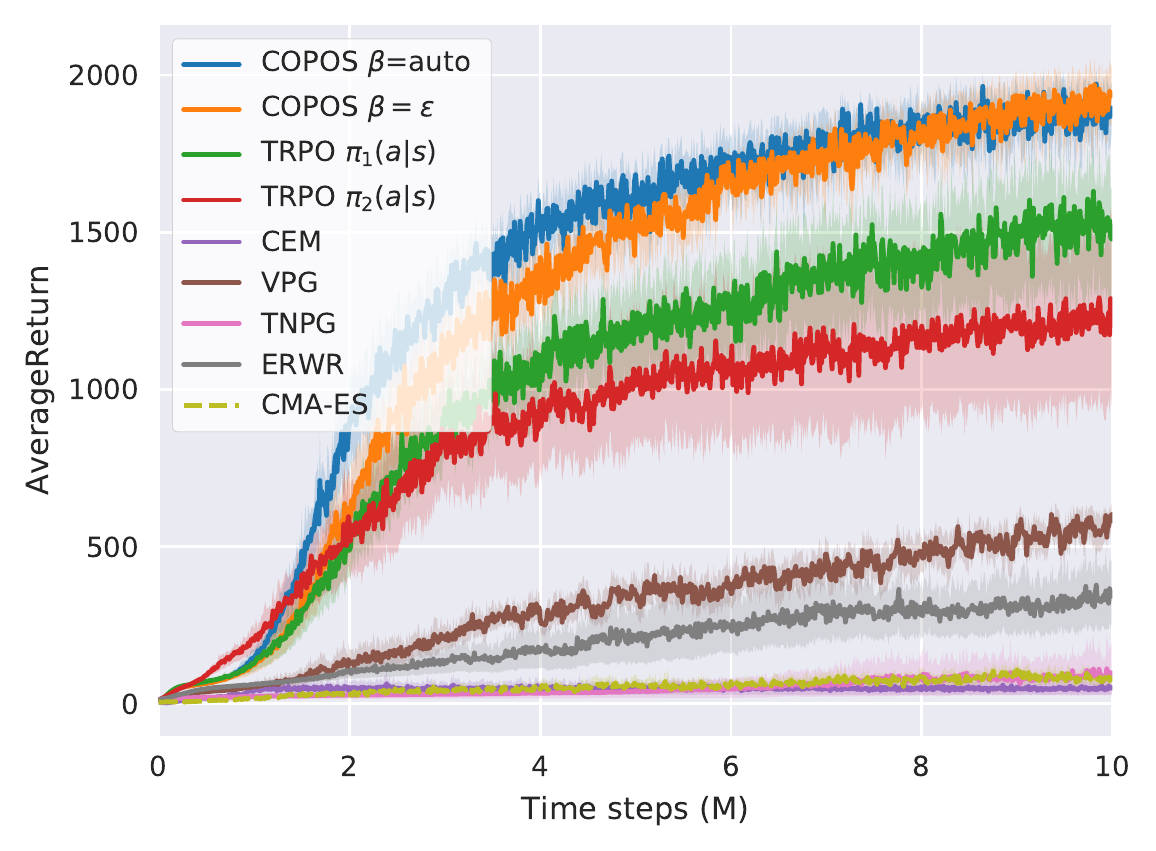}\vspace{-0.3em} &
    \includegraphics[width=0.5\textwidth]{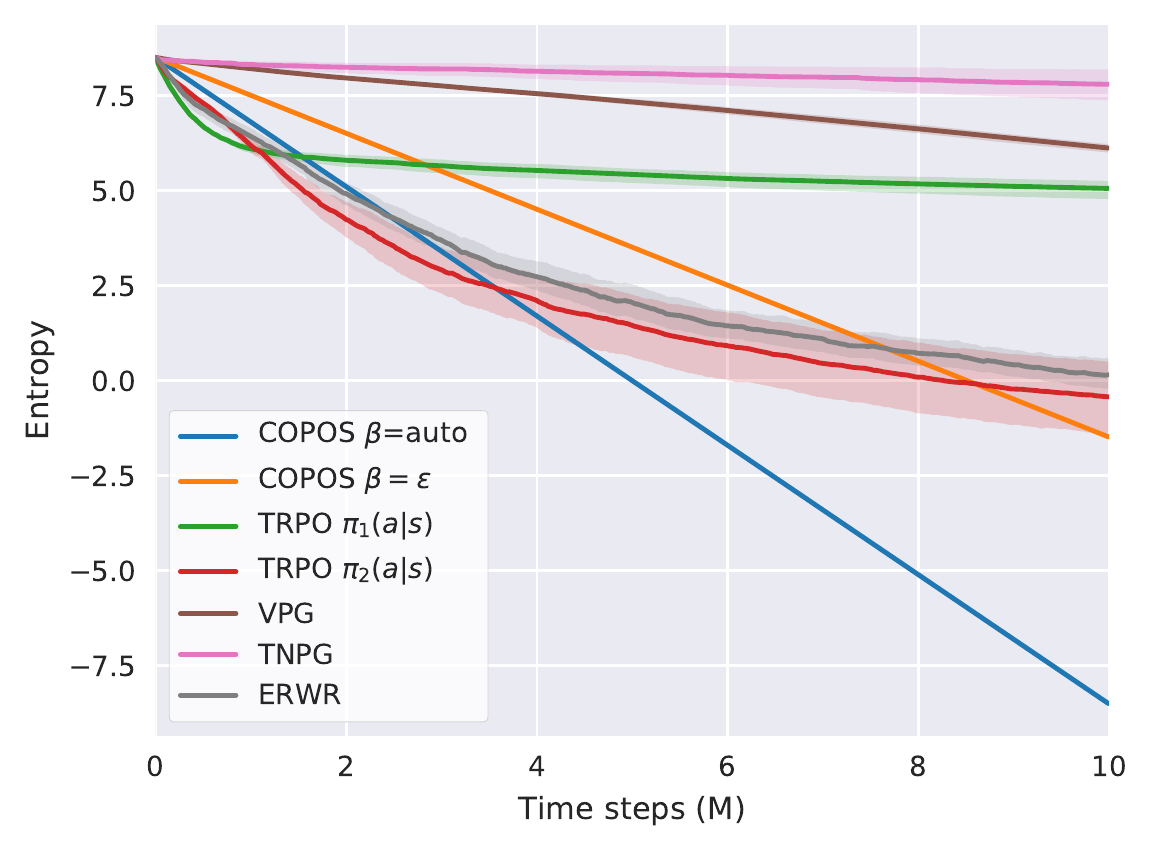}\vspace{-0.3em} \\
    \multicolumn{2}{c}{\scriptsize (c) RoboschoolWalker2d-v1} \\
    \includegraphics[width=0.5\textwidth]{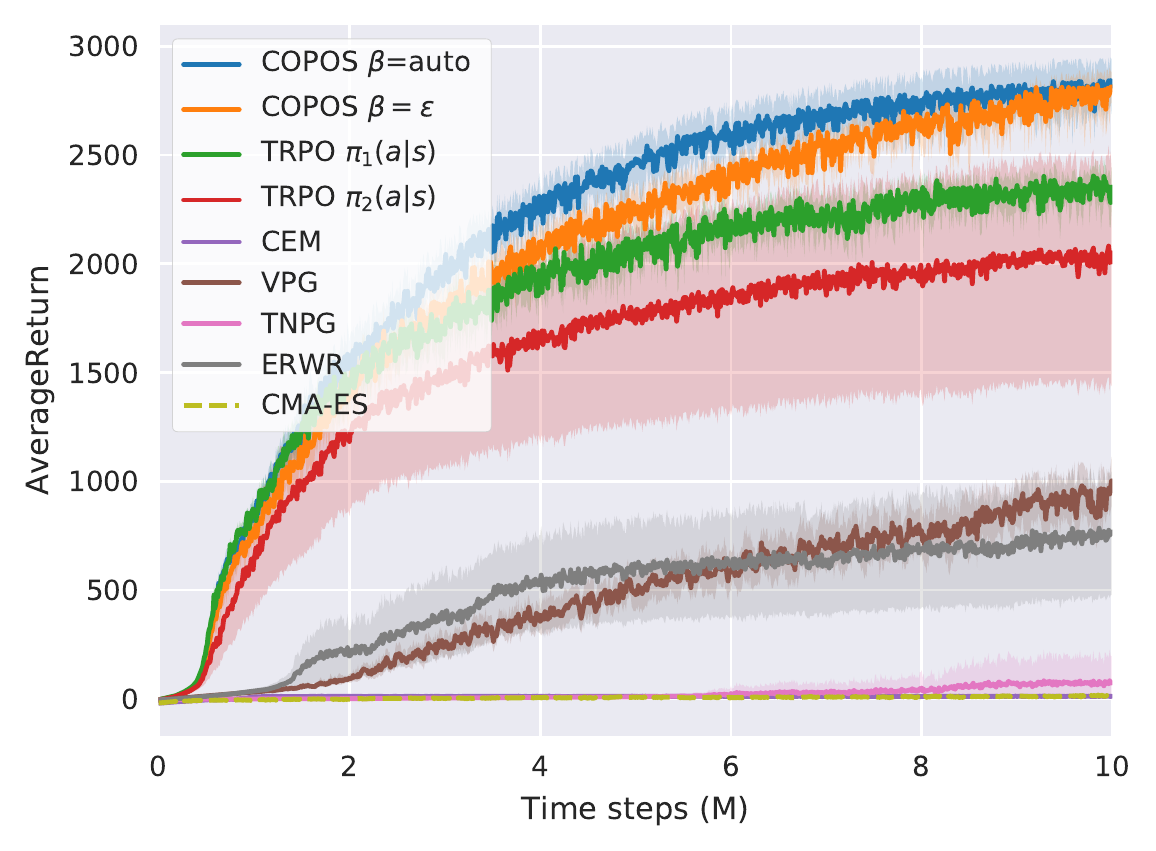}\vspace{-0.3em} &
    \includegraphics[width=0.5\textwidth]{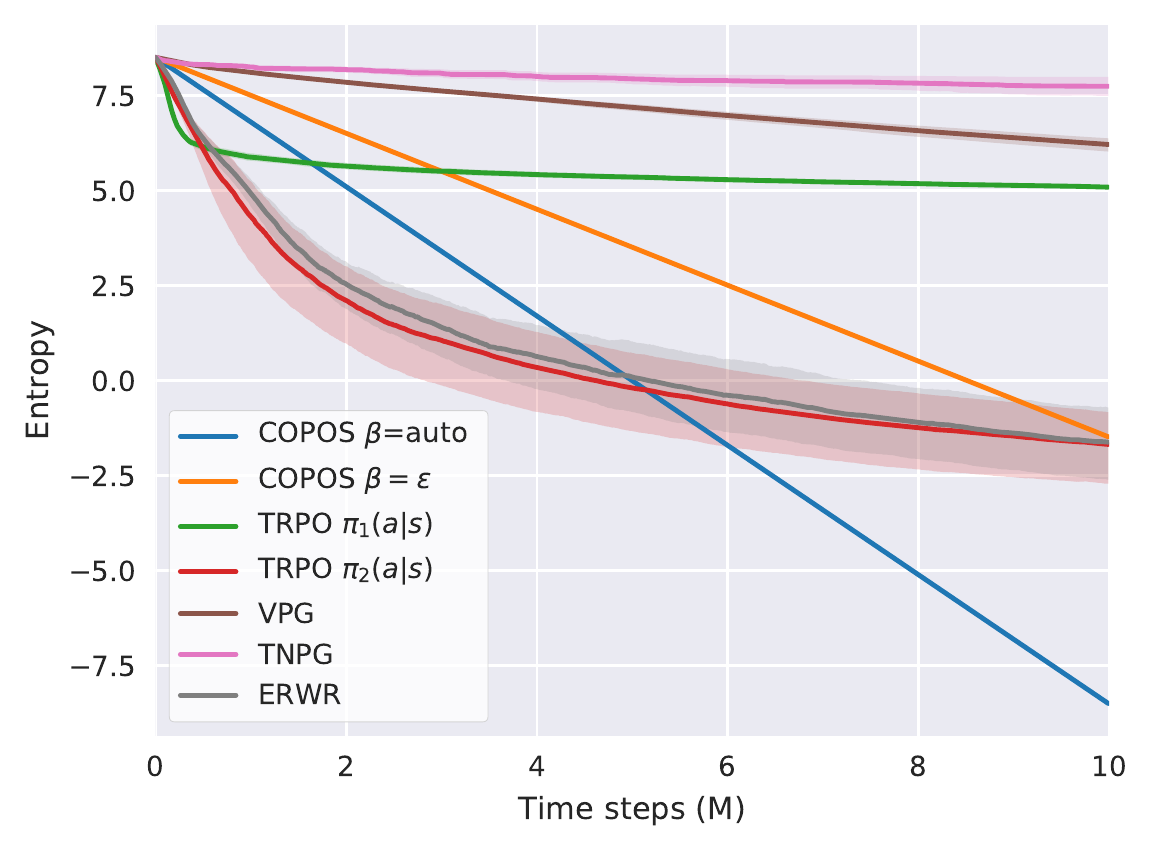}\vspace{-0.3em} \\
    \multicolumn{2}{c}{\scriptsize (d) RoboschoolHalfCheetah-v1}
  \end{tabular}
  \caption{\label{fig:continuous_problems1}Average return and differential entropy over $10$ random seeds of comparison methods in continuous Roboschool tasks (see, Fig.~\ref{fig:continuous_problems2} for the other continuous tasks and Table~\ref{tab:continuous} for a summary). Shaded area denotes the bootstrapped $95\%$ confidence interval. Algorithms were executed for $1000$ iterations with $10000$ time steps (samples) in each iteration.}
\end{figure}
\begin{figure}[t]
  \centering
  \tabcolsep=0.0cm
  \begin{tabular}{cc}	
    \includegraphics[width=0.5\textwidth]{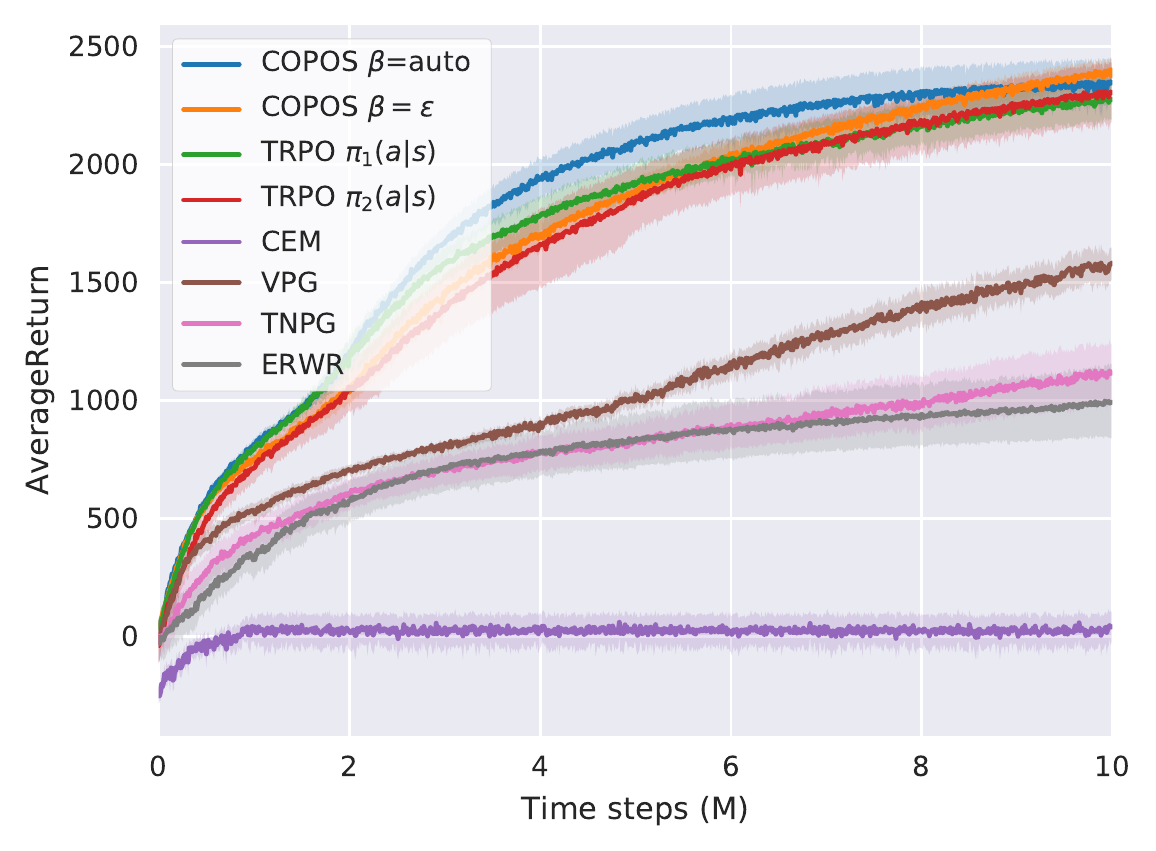}\vspace{-0.3em} &
    \includegraphics[width=0.5\textwidth]{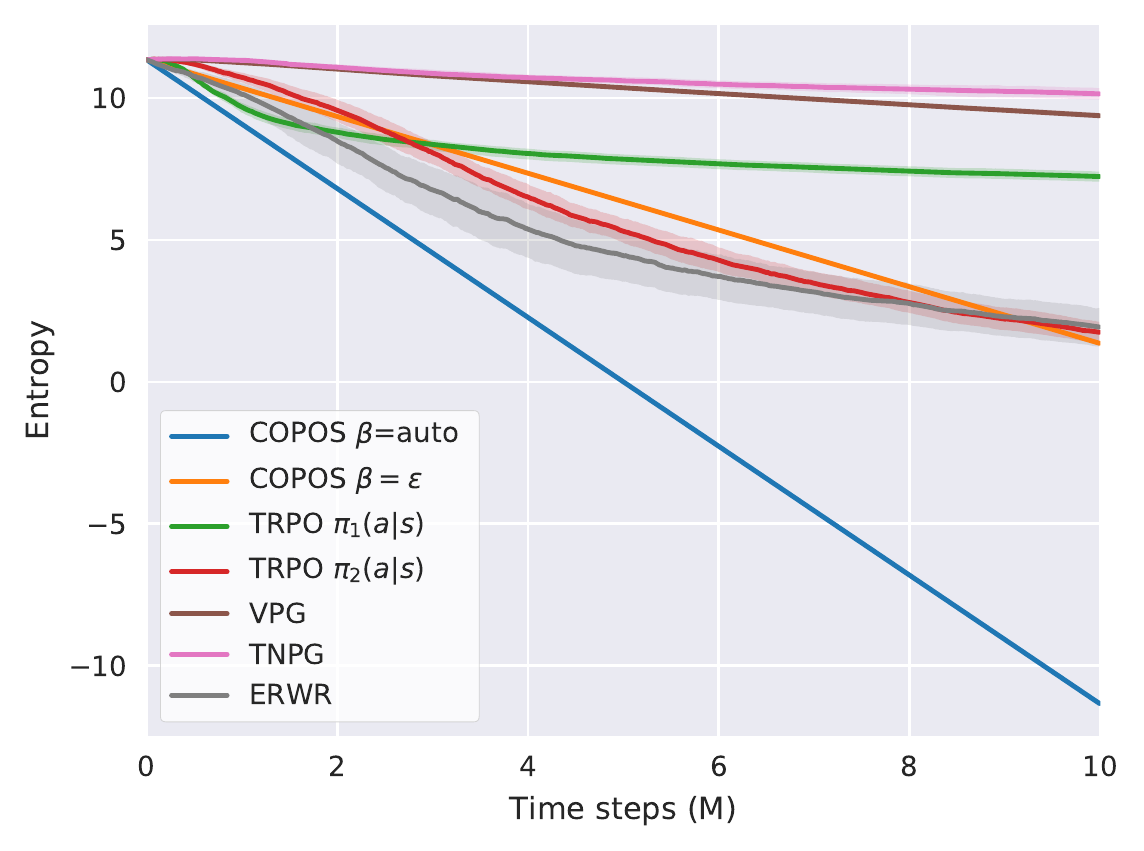}\vspace{-0.3em} \\
    \multicolumn{2}{c}{\scriptsize (a) RoboschoolAnt-v1} \\
    \includegraphics[width=0.5\textwidth]{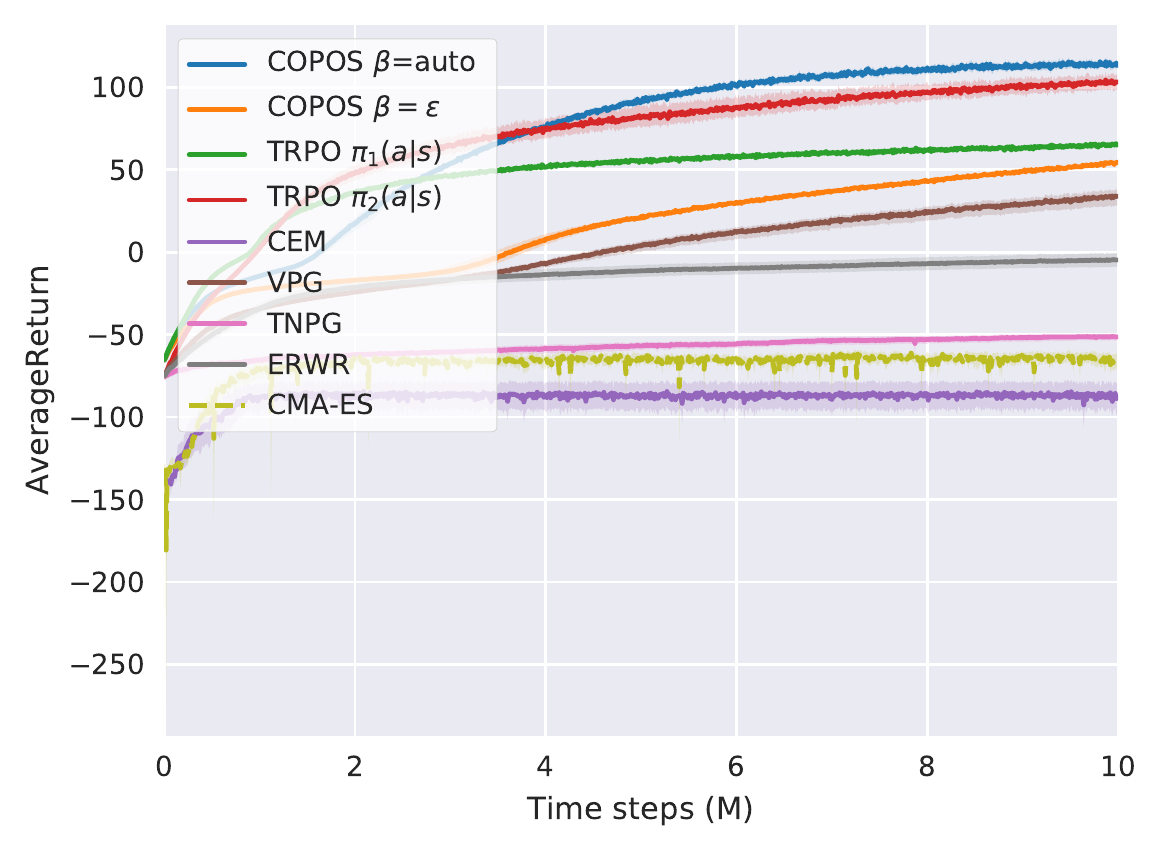}\vspace{-0.3em} &
    \includegraphics[width=0.5\textwidth]{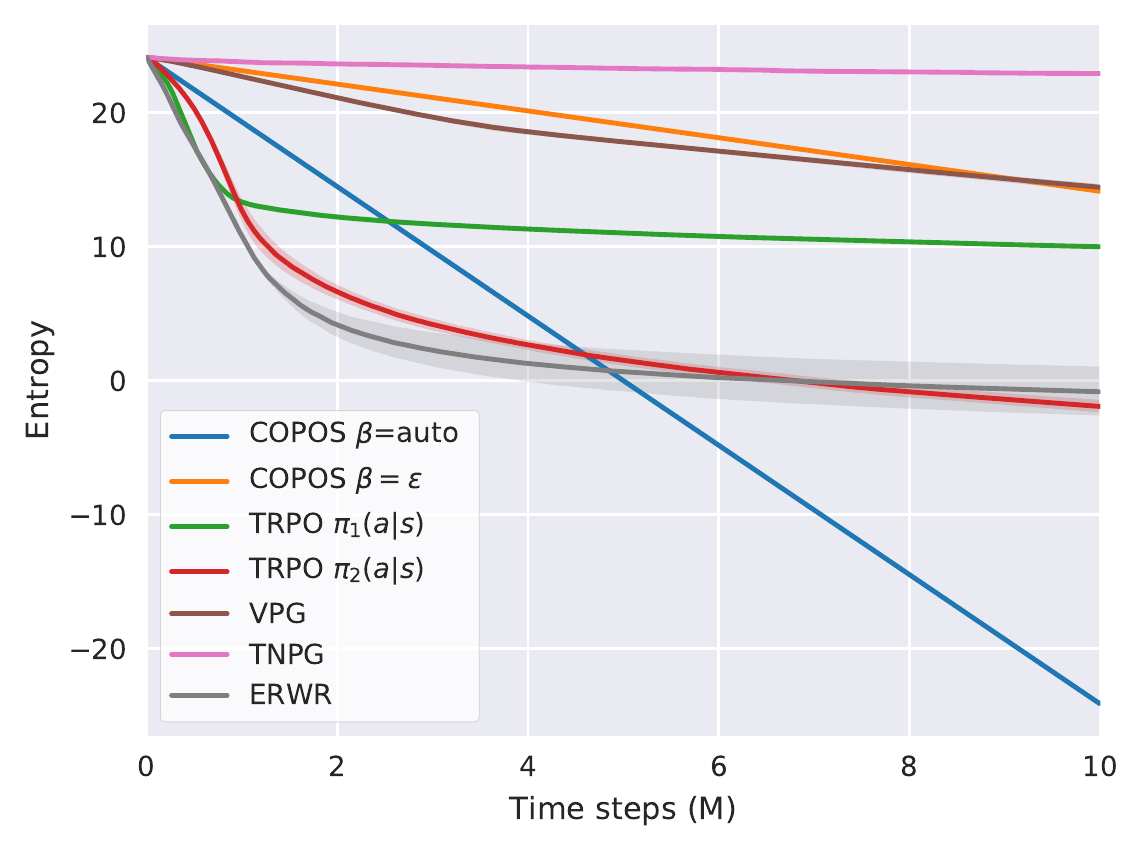}\vspace{-0.3em} \\
    \multicolumn{2}{c}{\scriptsize (b) RoboschoolHumanoid-v1} \\
    \includegraphics[width=0.5\textwidth]{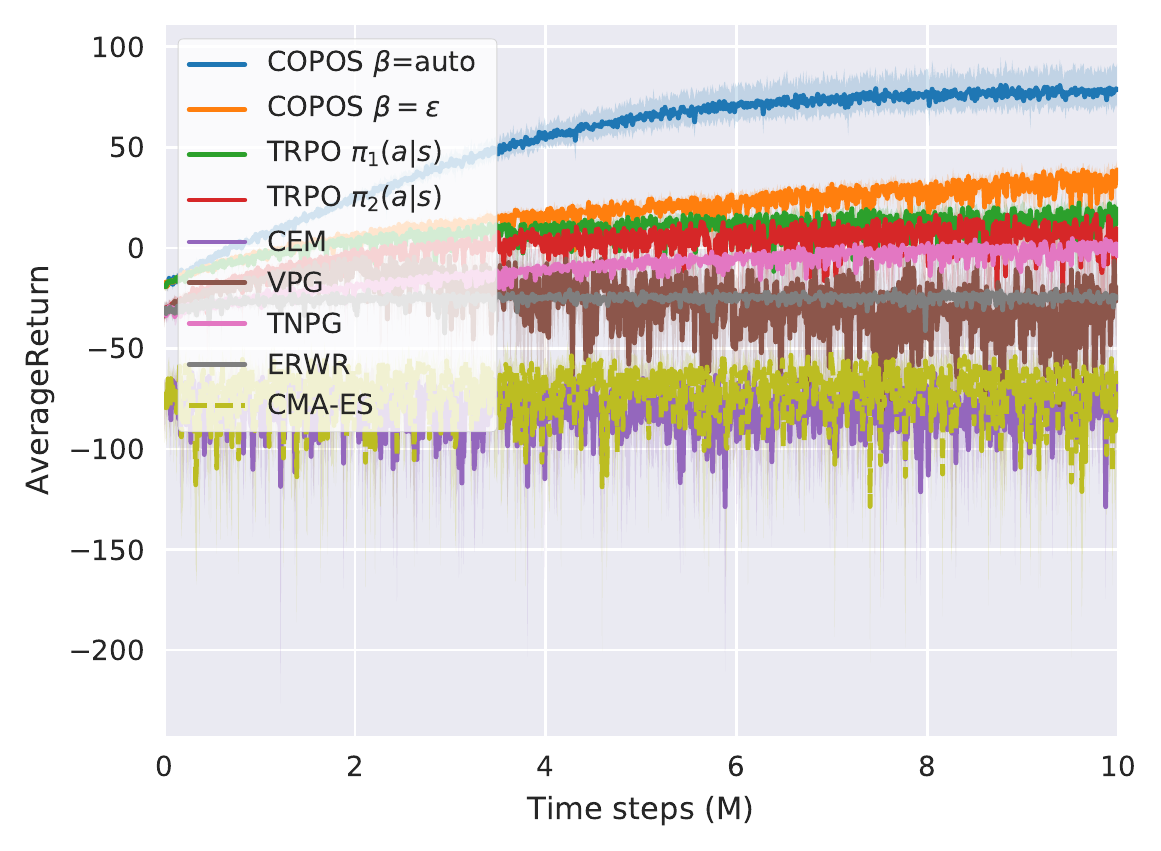}\vspace{-0.3em} &
    \includegraphics[width=0.5\textwidth]{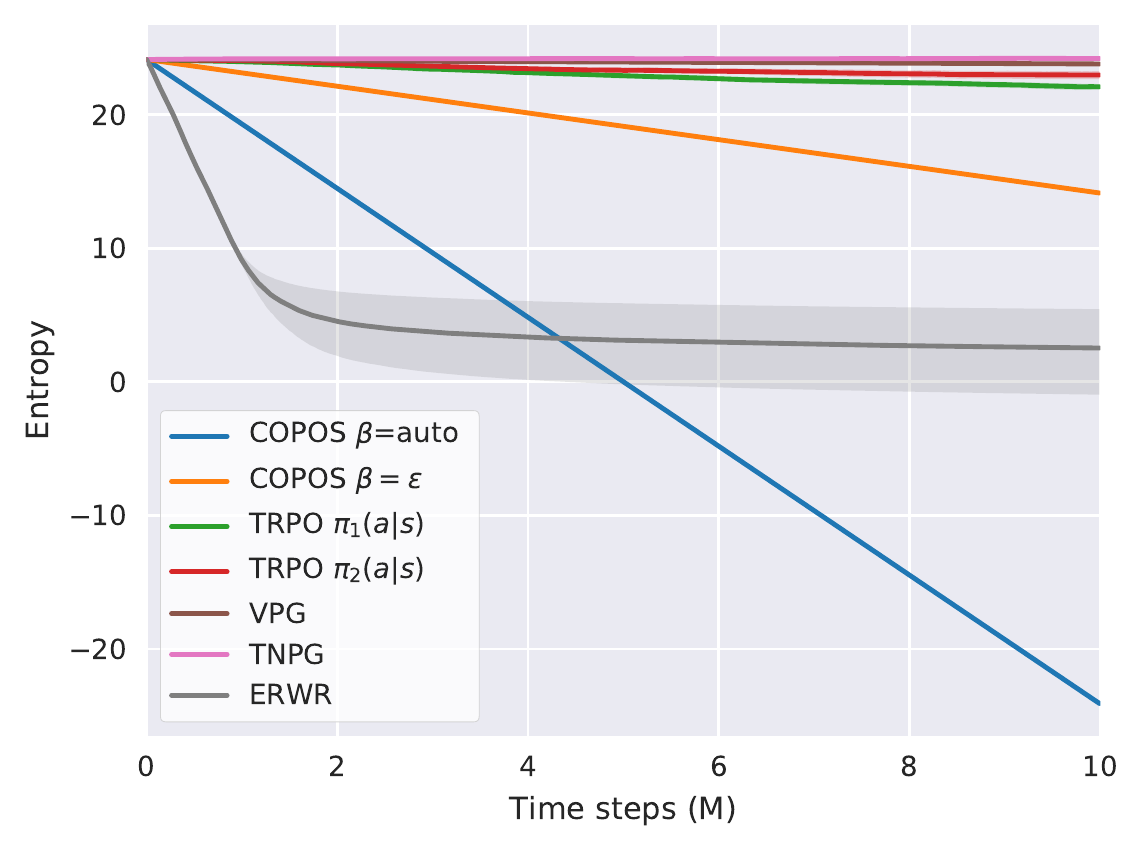}\vspace{-0.3em} \\
    \multicolumn{2}{c}{\scriptsize (c) RoboschoolHumanoidFlagrunHarder-v1} \\
    \includegraphics[width=0.5\textwidth]{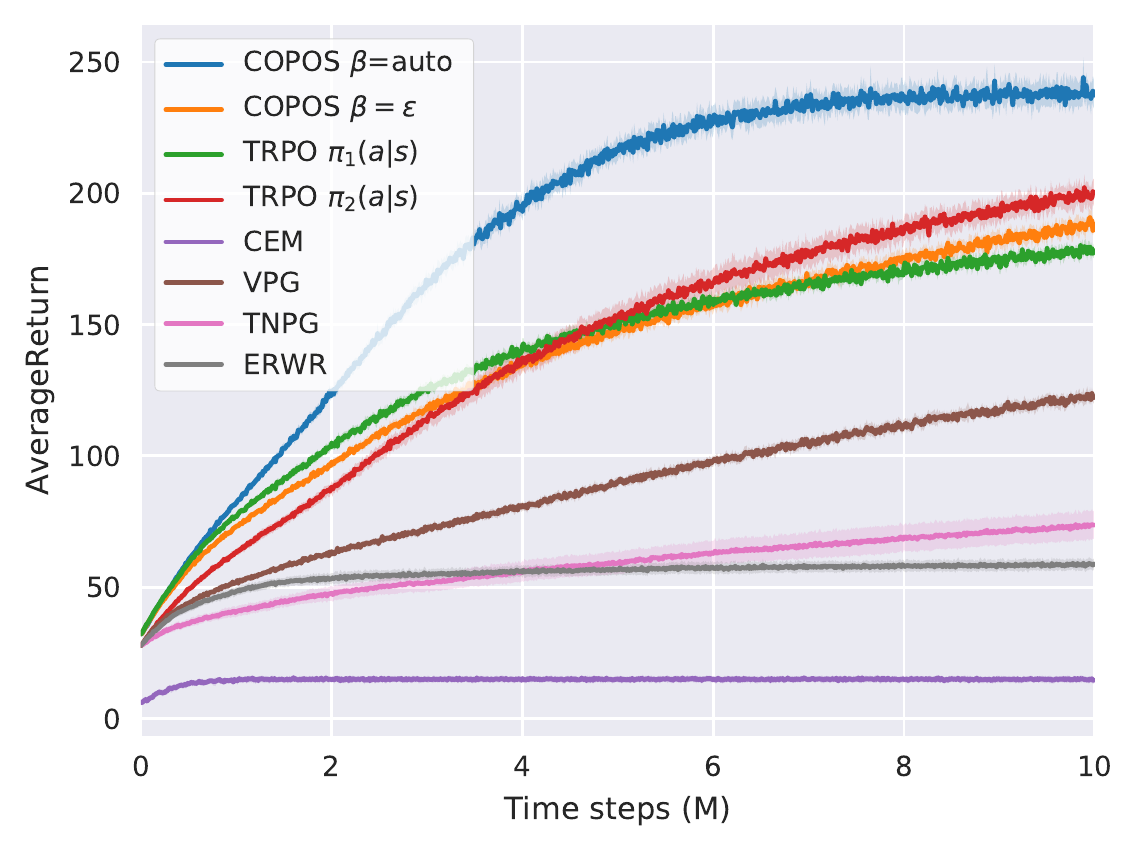}\vspace{-0.3em} &
    \includegraphics[width=0.5\textwidth]{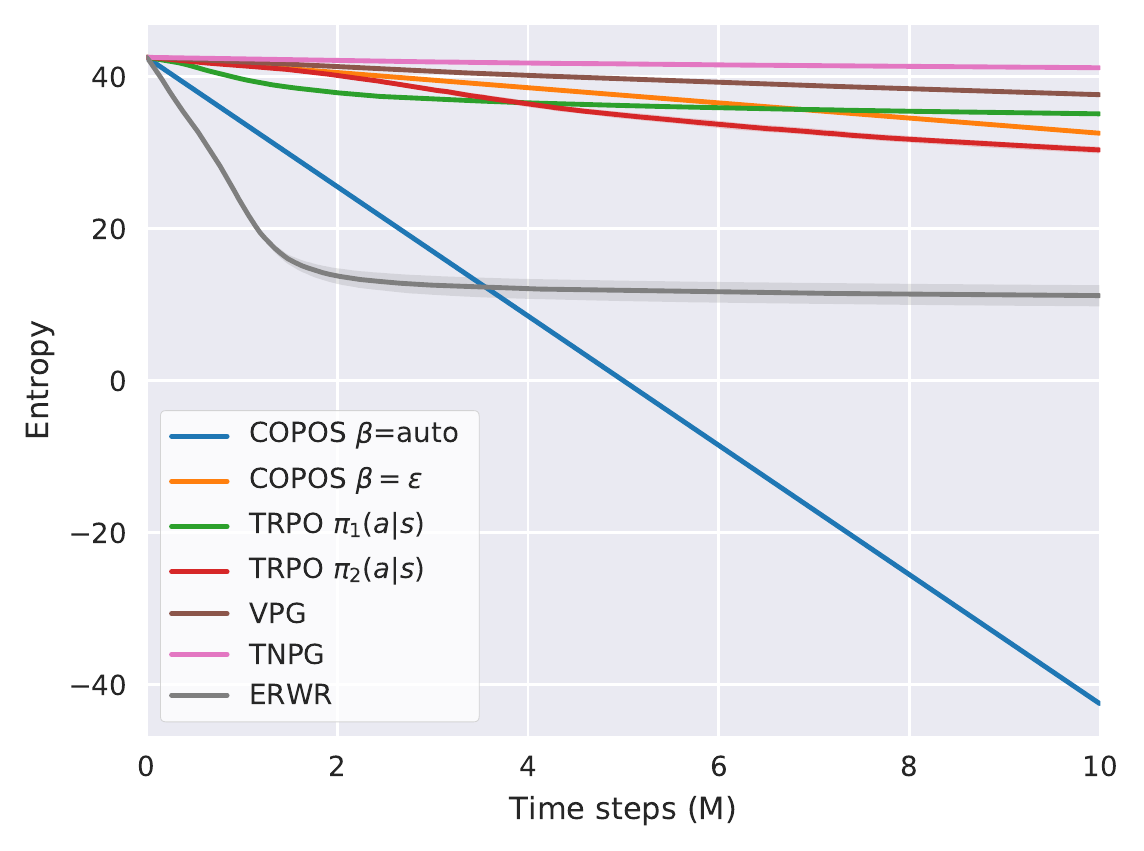}\vspace{-0.3em} \\
    \multicolumn{2}{c}{\scriptsize (d) RoboschoolAtlasForwardWalk-v1}
  \end{tabular}
  \caption{\label{fig:continuous_problems2}Average return and differential entropy of comparison methods over $10$ random seeds in continuous Roboschool tasks (see, Fig.~\ref{fig:continuous_problems1} for the other continuous tasks and Table~\ref{tab:continuous} for a summary). Shaded area denotes the bootstrapped $95\%$ confidence interval.
  Algorithms were executed for $1000$ iterations with $10000$ time steps (samples) in each iteration.}
\end{figure}

\begin{figure}[t]
  \centering
  \tabcolsep=0.0cm
  \begin{tabular}{cc}	
    \includegraphics[width=0.5\textwidth]{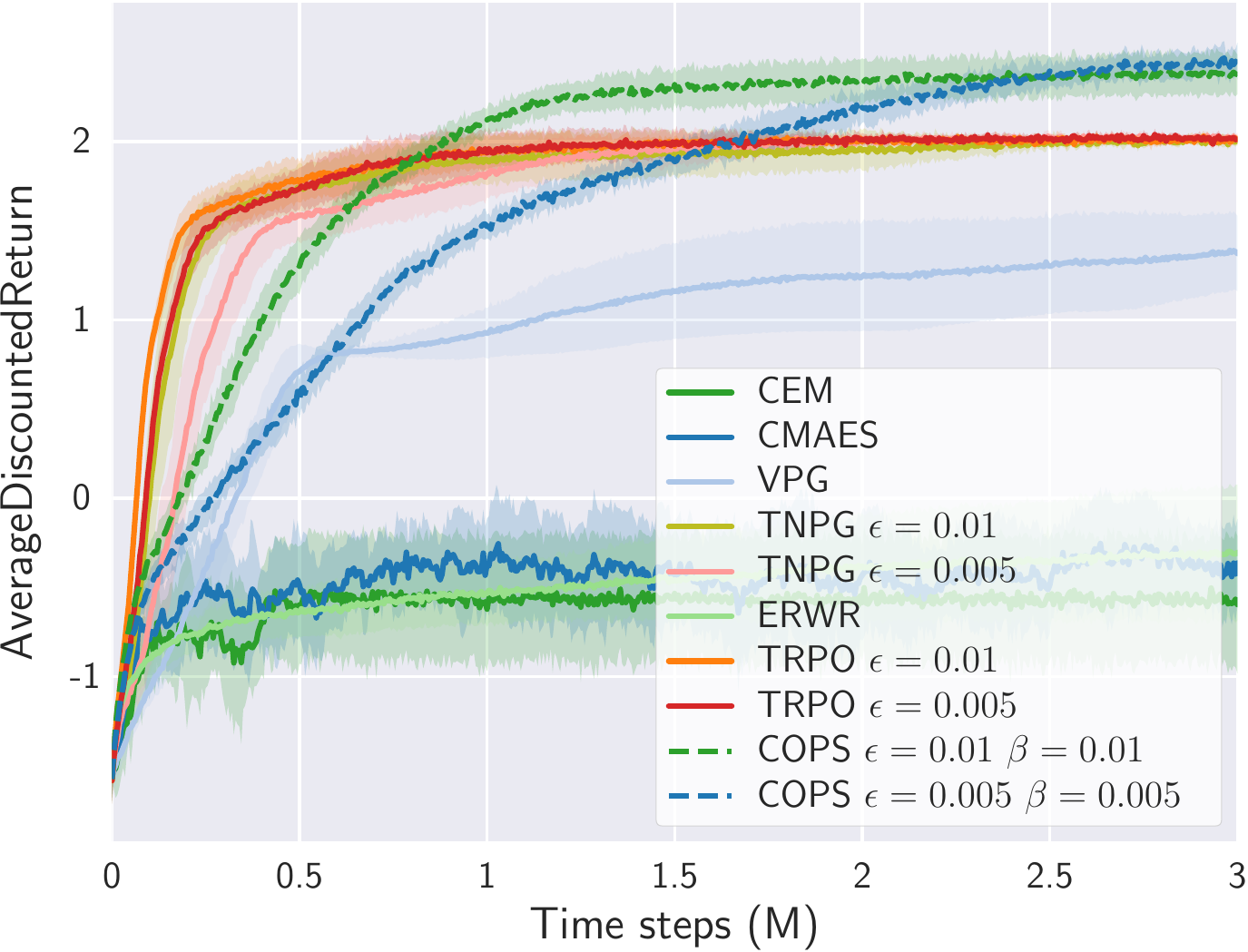}\vspace{-0.3em} &
    \includegraphics[width=0.5\textwidth]{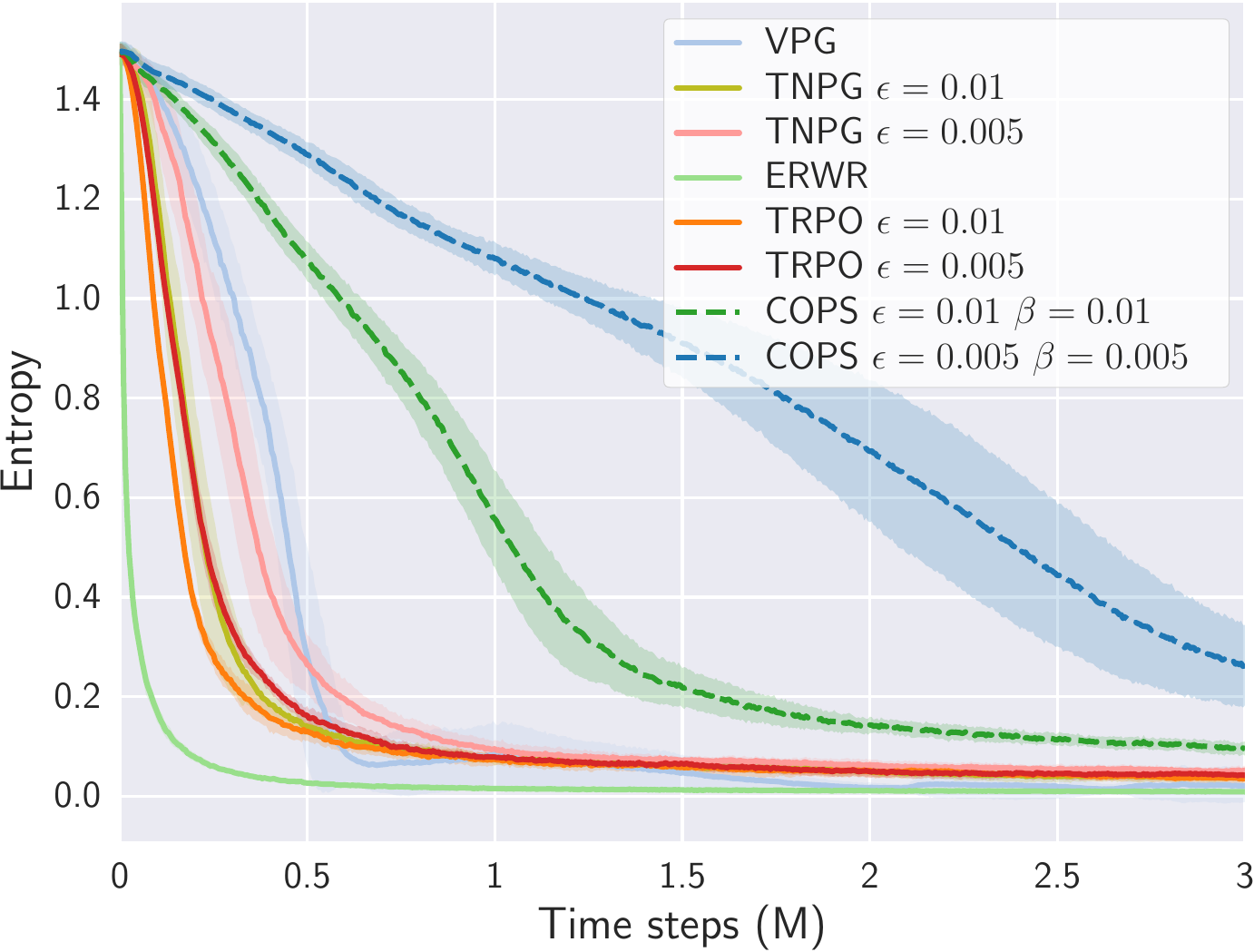}\vspace{-0.3em} \\
    \multicolumn{2}{c}{\scriptsize (a) FVRS $(5,7)$ noisy sensor} \\
    \includegraphics[width=0.5\textwidth]{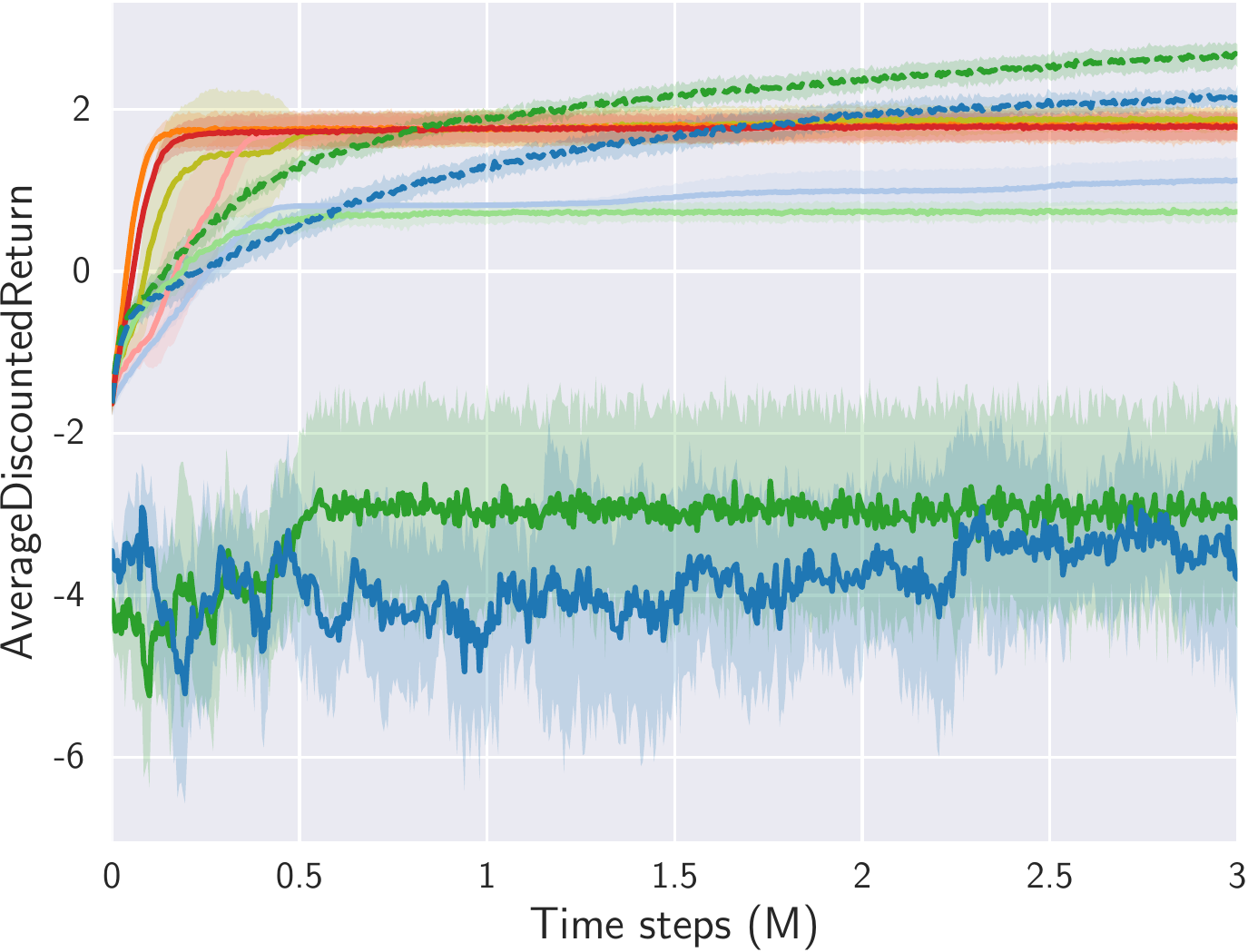}\vspace{-0.3em} &
    \includegraphics[width=0.5\textwidth]{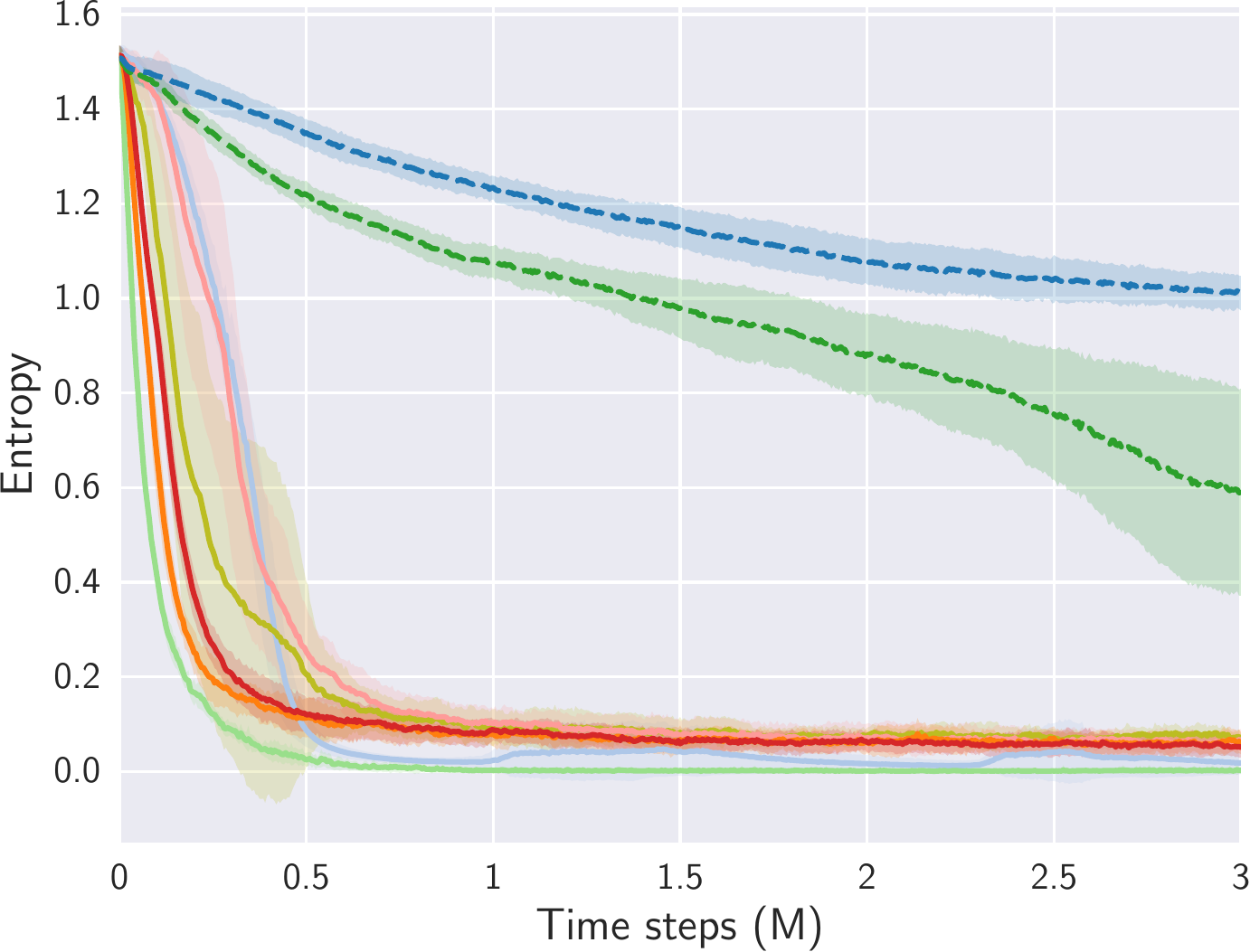}\vspace{-0.3em} \\
    \multicolumn{2}{c}{\scriptsize (b) FVRS $(5,7)$ full observations}
  \end{tabular}
  \caption{\label{fig:rocksample5x7_all_algos}Average discounted return and Shannon entropy for both FVRS
    $5 \times 7$ with a noisy sensor and full observations over $10$ random seeds. Shaded area denotes the bootstrapped $95\%$ confidence interval. Algorithms were executed for $600$ iterations with $5000$ time steps (samples) in each iteration.}
\end{figure}

\textbf{Continuous Tasks.}
In the continuous case, we ran experiments in eight different
Roboschool~\footnote{\url{https://github.com/openai/roboschool}}
environments which provide continuous
control tasks of increasing difficulty and action dimensions without
requiring a paid license. 

We ran all evaluations, $10$ random seeds for each method, for $1000$ iterations of
$10000$ samples each. In all problems, we used the Gaussian policy
defined in Eq.~(\ref{eq:gaussian_policy}) for \Acronym{} and TRPO (denoted
by $\pi_1(a|s)$) with $\max(10, \textrm{action dimensions})$ neural
network outputs as basis functions, a neural network with two hidden
layers each containing $32$ tanh-neurons, and a diagonal precision
matrix. For TRPO and other methods, except \Acronym{}, we also
evaluated a policy, denoted for TRPO by $\pi_2(a|s)$, where the neural
network directly specifies the mean and the diagonal covariance is
parameterized with log standard deviations. In the experiments, we used
high identical initial variances (we tried others without success) for
all policies. We set $\epsilon = 0.01$~\citep{schulman2015trust} in all experiments. For
\Acronym{} we used two equality entropy constraints: $\beta =
\epsilon$ and $\beta=$auto. In $\beta=$auto, we assume positive
initial entropy and schedule the entropy to be the negative initial
entropy after $1000$ iterations. Since we always initialize the
variances to one, higher dimensional problems have higher initial
entropy. Thus $\beta=$auto reduces the entropy faster for high
dimensional problems effectively scaling the reduction with
dimensionality.
Table~\ref{tab:continuous_problems} summarizes the results in
continuous tasks: \Acronym{} outperforms comparison methods in most of the
environments. Fig.~\ref{fig:continuous_problems1} and Fig.~\ref{fig:continuous_problems2} show learning curves and
performance of \Acronym{} compared to the other methods. \Acronym{} prevents both too fast, and, too slow entropy
reduction while outperforming comparison methods. Table~\ref{tab:continuous_TRPO_bonus} in Appendix~\ref{sec:TRPO_entropy_bonus} shows additional results for experiments where different constant entropy bonuses were added to the objective function of TRPO without success, highlighting
the necessity of principled entropy control.

\textbf{Discrete control task.}
Partial observability often requires efficient exploration due to
non-myopic actions yielding long term rewards which is challenging for
model-free methods. The Field Vision Rock Sample (FVRS)~\citep{ross08}
task is a partially observable Markov decision process (POMDP)
benchmark task. For the discrete action experiments we used as policy
a softmax policy with a fully connected feed forward neural network
consisting of 2 hidden layers with 30 tanh nonlinearities each. The
input to the neural network is the observation history and the current
position of the agent in the grid. To obtain the hyperparameters
$\beta$, $\epsilon$, and the scaling factor for TRPO with additive
entropy regularization, denoted with ``TRPO ent reg'', we performed a
grid search on smaller instances of FVRS. See Appendix~\ref{sec:technical_details_for_experiments} for more
details about the setup. Results in
Table~\ref{table:FVRS_comparison_algo} and
Figure~\ref{fig:rocksample5x7_all_algos} show that \Acronym{}
outperforms the comparison methods due to maintaining higher
entropy. FVRS has been used with model-based online POMDP
algorithms~\citep{ross08} but not with model-free algorithms. The best
model-based results in~\citep{ross08} (scaled to correspond to our
rewards, COPOS in parentheses) are $2.275$ ($1.94$) in FVRS(5,5) and
$2.34$ ($2.45$) in FVRS(5,7).

\section{Conclusions \& Future Work}
We showed that when we use the natural parameterization of a standard
exponential policy distribution in combination with compatible value
function approximation, the natural gradient and trust region
optimization are equivalent. Furthermore, we demonstrated that natural
gradient updates may reduce the entropy of the policy according to a
schedule which can lead to premature convergence. To combat the
problem of bad entropy scheduling in trust region methods we proposed
a new compatible policy search method called \Acronym{} that can
control the entropy of the policy using an entropy bound. In both
challenging high dimensional continuous and discrete tasks the
approach yielded state-of-the-art results due to better entropy control.
In future work, an exciting direction is to apply efficient
approximations to compute the natural gradient~\citep{bernacchia18}.
Moreover, we have started work on applying the proposed algorithm in challenging
partially observable environments found for example in autonomous driving
where exploration and sample efficiency is crucial for finding high quality policies~\citep{dosovitskiy2017carla}.

\begin{table}[h]
  \caption{\label{tab:continuous_problems} Continuous control environments. Mean of
    the average return over $50$ last iterations $\pm$ standard error
    over $10$ random seeds. Bold denotes: no statistically significant
    difference to the best result (Welch's t-test with $p <
    0.05$).}
  \label{tab:continuous}
  \begin{tabular}{lrrr}
    \hline\noalign{\smallskip}
    & \Acronym{} $\beta$=auto & \Acronym{} $\beta$=$\epsilon$ & TRPO $\pi_1(\vec{a}|\vec{s})$ \\	
    \noalign{\smallskip}\hline\noalign{\smallskip}
    RoboInvertedDoublePendulum-v1 & 7107  $\pm$  416 &  7722  $\pm$  235.3 &  7904  $\pm$  192.4 \\
    RoboHopper-v1 &  \textbf{2514  $\pm$  35.11} &  \textbf{2427  $\pm$  47.96} &  2005  $\pm$  26.04 \\
    RoboWalker2d-v1 &  \textbf{1878  $\pm$  34.59} &  \textbf{1906  $\pm$  41.48} &  1518  $\pm$  93.11 \\
    RoboHalfCheetah-v1 &  \textbf{2804  $\pm$  57.27} &  \textbf{2772  $\pm$  42.23} &  2341  $\pm$  16.52 \\
    RoboAnt-v1 &  \textbf{2335  $\pm$  49.69} &  \textbf{2375  $\pm$  26.88} &  \textbf{2263  $\pm$  48.09} \\
    RoboHumanoid-v1 & \textbf{113.85 $\pm$  0.94} &  52.92  $\pm$  0.31 &  64.77  $\pm$  0.38 \\
    RoboHumanoidFlagrunHarder-v1 &  \textbf{77.72  $\pm$  4.86} &  32.14  $\pm$  1.34 &  14.51  $\pm$  1.25 \\
    RoboAtlasForwardWalk-v1 &  \textbf{238.1  $\pm$  2.08} &  186.8  $\pm$  1.18 &  177.7  $\pm$  0.44 \\
    \noalign{\smallskip}\hline
  \end{tabular}
  \begin{tabular}{lrrr}
      \hline\noalign{\smallskip}
      & TRPO $\pi_2(\vec{a}|\vec{s})$ & CMA-ES & VPG \\	
      \noalign{\smallskip}\hline\noalign{\smallskip}
      RoboInvertedDoublePendulum-v1 &  8052  $\pm$  172.3 & 4382  $\pm$  379.2 & \textbf{9020  $\pm$ 35.10} \\
      RoboHopper-v1 &  2106  $\pm$  85.59 & 25.87  $\pm$  8.66 & 751.4  $\pm$  145.4 \\
      RoboWalker2d-v1 &  1223  $\pm$  139.0 & 82.38  $\pm$  9.72 &  559.7  $\pm$  13.55\\
      RoboHalfCheetah-v1 &  2024  $\pm$  261.2 & 13.38  $\pm$  3.88 &  918.1  $\pm$  43.57 \\
      RoboAnt-v1 &  \textbf{2291  $\pm$  65.66} &  Out of Memory &  1558  $\pm$  35.71 \\
      RoboHumanoid-v1 &  102.5  $\pm$  2.16 & -65.07  $\pm$  1.71 &  32.72  $\pm$  2.47 \\
      RoboHumanoidFlagrunHarder-v1 &  5.04  $\pm$  2.70 & -74.62  $\pm$  3.28 &  -31.62  $\pm$  4.59 \\
      RoboAtlasForwardWalk-v1 &  198.4  $\pm$  1.94 & Out of Memory & 121.5  $\pm$  0.92\\
      \noalign{\smallskip}\hline
  \end{tabular}
  \begin{tabular}{lrrr}
      \hline
      & CEM & TNPG & ERWR \\	
      \hline
      RoboInvertedDoublePendulum-v1 & 2643 $\pm$  628.9 & 4866  $\pm$  1178 &  6367  $\pm$  937.4 \\
      RoboHopper-v1 &  346.6  $\pm$  66.89 &  20.55  $\pm$  2.42 &  985.1  $\pm$  173.1 \\
      RoboWalker2d-v1 & 49.03  $\pm$  10.69 &  90.36  $\pm$  35.52 &  333.3  $\pm$  47.73 \\
      RoboHalfCheetah-v1 & 11.27  $\pm$  2.93 & 72.22  $\pm$  56.48 &  747.3  $\pm$  137.8 \\
      RoboAnt-v1 & 25.17  $\pm$  35.09 &  1102  $\pm$  64.61 &  982.6  $\pm$  74.23 \\
      RoboHumanoid-v1 & -87.19  $\pm$  4.21 &  -51.39  $\pm$  1.01 &  -4.95  $\pm$  2.16 \\
      RoboHumanoidFlagrunHarder-v1 & -78.00  $\pm$  4.74 &  -0.25  $\pm$  0.68 &  -24.59  $\pm$  0.65\\
      RoboAtlasForwardWalk-v1 & 14.90  $\pm$  0.60 &  72.96  $\pm$  2.75 &  58.54  $\pm$  1.07 \\
      \hline 
  \end{tabular}
\end{table}

\begin{table}[h]
  \caption{\label{table:FVRS_comparison_algo}Average discounted return on discrete control FVRS instances (fully and partially observable). Bold denotes: no statistically significant difference to the best result (Welch's t-test with $p < 0.05$).}
  \label{tab:FVRS}
  \begin{tabular}{lcccc}
    \hline\noalign{\smallskip}
    & \Acronym{} & TRPO & TRPO ent reg & TNPG \\
    \noalign{\smallskip}\hline\noalign{\smallskip}	
    $5 \times 5$ full & \textbf{2.14} $\pm$ \textbf{0.08} & 1.50 $\pm$ 0.23 & 1.47 $\pm$ 0.22 & 1.43 $\pm$ 0.28 \\
    $5 \times 5$ noise & \textbf{1.94} $\pm$ \textbf{0.12} & 1.24 $\pm$ 0.02 & 1.24 $\pm$ 0.01 & 1.24 $\pm$ 0.01 \\
    $5 \times 7$ full & \textbf{2.66} $\pm$ \textbf{0.14} & 1.80 $\pm$ 0.20 & 1.92 $\pm$ 0.17 & 1.87 $\pm$ 0.16 \\
    $5 \times 7$ noise & \textbf{2.45} $\pm$ \textbf{0.10} & 2.01 $\pm$ 0.02 & 2.02 $\pm$ 0.02 & 2.01 $\pm$ 0.03 \\
    $7 \times 8$ full & \textbf{1.66} $\pm$ \textbf{0.17} & 1.28 $\pm$ 0.32 & 1.31 $\pm$ 0.20 & 1.19 $\pm$ 0.26 \\
    $7 \times 8$ noise & \textbf{1.32} $\pm$ \textbf{0.15} & \textbf{1.22} $\pm$ \textbf{0.28} & \textbf{1.36} $\pm$ \textbf{0.16} & \textbf{1.18} $\pm$ \textbf{0.22}\\
    \noalign{\smallskip}\hline
  \end{tabular}
\end{table}

\section*{Acknowledgements}
  This work was supported by EU Horizon 2020 project RoMaNS and ERC StG SKILLS4ROBOTS, project references \#645582 and \#640554, and, by German Research Foundation project PA 3179/1-1 (ROBOLEAP).%

\FloatBarrier
  
\bibliographystyle{plainnat}
\bibliography{copos.bib}

\begin{thebibliography}{31}
\providecommand{\natexlab}[1]{#1}
\providecommand{\url}[1]{\texttt{#1}}
\expandafter\ifx\csname urlstyle\endcsname\relax
  \providecommand{\doi}[1]{doi: #1}\else
  \providecommand{\doi}{doi: \begingroup \urlstyle{rm}\Url}\fi

\bibitem[Abdolmaleki et~al.(2015)Abdolmaleki, Lioutikov, Peters, Lau, Reis, and
  Neumann]{Abdolmaleki_NIPS2015}
A.~Abdolmaleki, R.~Lioutikov, J~Peters, N.~Lau, L.~Reis, and G.~Neumann.
\newblock Model-based relative entropy stochastic search.
\newblock In \emph{Advances in Neural Information Processing Systems (NIPS)}.
  mit press, 2015.

\bibitem[Abdolmaleki et~al.(2018)Abdolmaleki, Springenberg, Tassa, Munos,
  Heess, and Riedmiller]{abdolmaleki18}
Abbas Abdolmaleki, Jost~T. Springenberg, Yuval Tassa, Remi Munos, Nicolas
  Heess, and Martin Riedmiller.
\newblock {Maximum a Posteriori Policy Optimisation}.
\newblock In \emph{Proceedings of the International Conference on Learning
  Representations (ICLR)}, 2018.

\bibitem[Akrour et~al.(2016)Akrour, Abdolmaleki, Abdulsamad, and
  Neumann]{akrour2016model-free}
R.~Akrour, A.~Abdolmaleki, H.~Abdulsamad, and G.~Neumann.
\newblock Model-free trajectory optimization for reinforcement learning.
\newblock In \emph{Proceedings of the International Conference on Machine
  Learning (ICML)}, 2016.

\bibitem[Akrour et~al.(2018)Akrour, Abdolmaleki, Abdulsamad, Peters, and
  Neumann]{akrour2018model}
Riad Akrour, Abbas Abdolmaleki, Hany Abdulsamad, Jan Peters, and Gerhard
  Neumann.
\newblock {Model-Free Trajectory-based Policy Optimization with Monotonic
  Improvement}.
\newblock \emph{Journal of Machine Learning Research}, 19\penalty0
  (14):\penalty0 1--25, 2018.

\bibitem[Amari(1998)]{amari98natural}
S.~Amari.
\newblock Natural gradient works efficiently in learning.
\newblock \emph{Neural Computation}, 10\penalty0 (2):\penalty0 251--276, 1998.

\bibitem[Bagnell and Schneider(2003)]{bagnell2003covariant}
J~Andrew Bagnell and Jeff Schneider.
\newblock Covariant policy search.
\newblock IJCAI, 2003.

\bibitem[Bernacchia et~al.(2018)Bernacchia, Lengyel, and
  Hennequin]{bernacchia18}
Alberto Bernacchia, Mate Lengyel, and Guillaume Hennequin.
\newblock Exact natural gradient in deep linear networks and its application to
  the nonlinear case.
\newblock In \emph{Advances in Neural Information Processing Systems (NIPS)},
  pages 5945--5954. Curran Associates, Inc., 2018.

\bibitem[Boyd and Vandenberghe(2004)]{boyd2004convex}
Stephen Boyd and Lieven Vandenberghe.
\newblock \emph{Convex optimization}.
\newblock Cambridge university press, 2004.

\bibitem[Daniel et~al.(2016)Daniel, Neumann, Kroemer, and
  Peters]{Daniel2016JMLR}
C.~Daniel, G.~Neumann, O.~Kroemer, and J.~Peters.
\newblock Hierarchical relative entropy policy search.
\newblock \emph{Journal of Machine Learning Research (JMLR)}, pages 1--50,
  2016.

\bibitem[Dosovitskiy et~al.(2017)Dosovitskiy, Ros, Codevilla, Lopez, and
  Koltun]{dosovitskiy2017carla}
Alexey Dosovitskiy, German Ros, Felipe Codevilla, Antonio Lopez, and Vladlen
  Koltun.
\newblock {CARLA: An Open Urban Driving Simulator}.
\newblock In \emph{Conference on Robot Learning}, pages 1--16, 2017.

\bibitem[Duan et~al.(2016)Duan, Chen, Houthooft, Schulman, and
  Abbeel]{DuanCHSA16}
Yan Duan, Xi~Chen, Rein Houthooft, John Schulman, and Pieter Abbeel.
\newblock Benchmarking deep reinforcement learning for continuous control.
\newblock In \emph{Proceedings of the 33nd International Conference on Machine
  Learning, {ICML} 2016, New York City, NY, USA, June 19-24, 2016}, pages
  1329--1338, 2016.
\newblock URL \url{http://jmlr.org/proceedings/papers/v48/duan16.html}.

\bibitem[Geist and Pietquin(2010)]{geist10}
Matthieu Geist and Olivier Pietquin.
\newblock Revisiting natural actor-critics with value function approximation.
\newblock In \emph{International Conference on Modeling Decisions for
  Artificial Intelligence}, pages 207--218. Springer, 2010.

\bibitem[Hansen and Ostermeier(2001)]{hansen2001completely}
Nikolaus Hansen and Andreas Ostermeier.
\newblock Completely derandomized self-adaptation in evolution strategies.
\newblock \emph{Evolutionary computation}, 9\penalty0 (2):\penalty0 159--195,
  2001.

\bibitem[Kakade(2001)]{Kakade:2001}
Sham Kakade.
\newblock A natural policy gradient.
\newblock In Thomas~G. Dietterich, Suzanna Becker, and Zoubin Ghahramani,
  editors, \emph{Advances in Neural Information Processing Systems 14 (NIPS
  2001)}, pages 1531--1538. MIT Press, 2001.

\bibitem[Kober and Peters(2009)]{NIPS2008_3545}
Jens Kober and Jan~R. Peters.
\newblock Policy search for motor primitives in robotics.
\newblock In D.~Koller, D.~Schuurmans, Y.~Bengio, and L.~Bottou, editors,
  \emph{Advances in Neural Information Processing Systems 21}, pages 849--856.
  Curran Associates, Inc., 2009.

\bibitem[Lillicrap et~al.(2015)Lillicrap, Hunt, Pritzel, Heess, Erez, Tassa,
  Silver, and Wierstra]{DBLP:journals/corr/LillicrapHPHETS15}
Timothy~P. Lillicrap, Jonathan~J. Hunt, Alexander Pritzel, Nicolas Heess, Tom
  Erez, Yuval Tassa, David Silver, and Daan Wierstra.
\newblock Continuous control with deep reinforcement learning.
\newblock \emph{CoRR}, abs/1509.02971, 2015.
\newblock URL \url{http://arxiv.org/abs/1509.02971}.

\bibitem[Mnih et~al.(2015)Mnih, Kavukcuoglu, Silver, Rusu, Veness, Bellemare,
  Graves, Riedmiller, Fidjeland, Ostrovski, Petersen, Beattie, Sadik,
  Antonoglou, King, Kumaran, Wierstra, Legg, and Hassabis]{mnih2015humanlevel}
Volodymyr Mnih, Koray Kavukcuoglu, David Silver, Andrei~A. Rusu, Joel Veness,
  Marc~G. Bellemare, Alex Graves, Martin Riedmiller, Andreas~K. Fidjeland,
  Georg Ostrovski, Stig Petersen, Charles Beattie, Amir Sadik, Ioannis
  Antonoglou, Helen King, Dharshan Kumaran, Daan Wierstra, Shane Legg, and
  Demis Hassabis.
\newblock Human-level control through deep reinforcement learning.
\newblock \emph{Nature}, 518\penalty0 (7540):\penalty0 529--533, February 2015.
\newblock ISSN 00280836.

\bibitem[Mnih et~al.(2016)Mnih, Badia, Mirza, Graves, Lillicrap, Harley,
  Silver, and Kavukcuoglu]{mnih2016asynchronous}
Volodymyr Mnih, Adria~Puigdomenech Badia, Mehdi Mirza, Alex Graves, Timothy
  Lillicrap, Tim Harley, David Silver, and Koray Kavukcuoglu.
\newblock Asynchronous methods for deep reinforcement learning.
\newblock In \emph{International Conference on Machine Learning}, pages
  1928--1937, 2016.

\bibitem[O'Donoghue et~al.(2016)O'Donoghue, Munos, Kavukcuoglu, and
  Mnih]{DBLP:journals/corr/ODonoghueMKM16}
Brendan O'Donoghue, R{\'{e}}mi Munos, Koray Kavukcuoglu, and Volodymyr Mnih.
\newblock {PGQ:} combining policy gradient and q-learning.
\newblock \emph{CoRR}, abs/1611.01626, 2016.
\newblock URL \url{http://arxiv.org/abs/1611.01626}.

\bibitem[Peters and Schaal(2008)]{4863}
J.~Peters and S.~Schaal.
\newblock Natural actor-critic.
\newblock \emph{Neurocomputing}, 71\penalty0 (7-9):\penalty0 1180--1190, March
  2008.

\bibitem[Peters et~al.(2010)Peters, M{\"u}lling, and Altun]{peters2010relative}
Jan Peters, Katharina M{\"u}lling, and Yasemin Altun.
\newblock Relative entropy policy search.
\newblock In \emph{AAAI}, pages 1607--1612. Atlanta, 2010.

\bibitem[Ross et~al.(2008)Ross, Pineau, Paquet, and Chaib-Draa]{ross08}
St{\'e}phane Ross, Joelle Pineau, S{\'e}bastien Paquet, and Brahim Chaib-Draa.
\newblock Online planning algorithms for {POMDPs}.
\newblock \emph{Journal of Artificial Intelligence Research}, 32:\penalty0
  663--704, 2008.

\bibitem[Rubinstein(1999)]{Rubinstein1999}
Reuven Rubinstein.
\newblock The cross-entropy method for combinatorial and continuous
  optimization.
\newblock \emph{Methodology And Computing In Applied Probability}, 1\penalty0
  (2):\penalty0 127--190, Sep 1999.

\bibitem[Schulman et~al.(2015)Schulman, Levine, Abbeel, Jordan, and
  Moritz]{schulman2015trust}
John Schulman, Sergey Levine, Pieter Abbeel, Michael Jordan, and Philipp
  Moritz.
\newblock Trust region policy optimization.
\newblock In \emph{Proceedings of the 32nd International Conference on Machine
  Learning (ICML-15)}, pages 1889--1897, 2015.

\bibitem[Schulman et~al.(2017)Schulman, Wolski, Dhariwal, Radford, and
  Klimov]{DBLP:journals/corr/SchulmanWDRK17}
John Schulman, Filip Wolski, Prafulla Dhariwal, Alec Radford, and Oleg Klimov.
\newblock Proximal policy optimization algorithms.
\newblock \emph{CoRR}, abs/1707.06347, 2017.
\newblock URL \url{http://arxiv.org/abs/1707.06347}.

\bibitem[Silver et~al.(2014)Silver, Lever, Heess, Degris, Wierstra, and
  Riedmiller]{silver2014deterministic}
David Silver, Guy Lever, Nicolas Heess, Thomas Degris, Daan Wierstra, and
  Martin Riedmiller.
\newblock Deterministic policy gradient algorithms.
\newblock In \emph{ICML}, 2014.

\bibitem[Sutton et~al.(1999)Sutton, McAllester, Singh, and
  Mansour]{Sutton:1999:PGM:3009657.3009806}
Richard~S. Sutton, David McAllester, Satinder Singh, and Yishay Mansour.
\newblock Policy gradient methods for reinforcement learning with function
  approximation.
\newblock In \emph{Proceedings of the 12th International Conference on Neural
  Information Processing Systems}, NIPS'99, pages 1057--1063, Cambridge, MA,
  USA, 1999. MIT Press.

\bibitem[Tangkaratt et~al.(2018)Tangkaratt, Abdolmaleki, and
  Sugiyama]{tangkaratt18}
Voot Tangkaratt, Abbas Abdolmaleki, and Masashi Sugiyama.
\newblock {Guide Actor-Critic for Continuous Control}.
\newblock In \emph{Proceedings of the International Conference on Learning
  Representations (ICLR)}, 2018.

\bibitem[Wierstra et~al.(2008)Wierstra, Schaul, Peters, and
  Schmidhuber]{wierstraCEC2008}
D.~Wierstra, T.~Schaul, J.~Peters, and J.~Schmidhuber.
\newblock Natural evolution strategies.
\newblock In \emph{IEEE Congress on Evolutionary Computation}, pages
  3381--3387. IEEE, 2008.

\bibitem[Williams(1992)]{williams1992simple}
Ronald~J Williams.
\newblock Simple statistical gradient-following algorithms for connectionist
  reinforcement learning.
\newblock \emph{Machine learning}, 8\penalty0 (3-4):\penalty0 229--256, 1992.

\bibitem[Wu et~al.(2017)Wu, Mansimov, Grosse, Liao, and Ba]{wu2017scalable}
Yuhuai Wu, Elman Mansimov, Roger~B Grosse, Shun Liao, and Jimmy Ba.
\newblock Scalable trust-region method for deep reinforcement learning using
  kronecker-factored approximation.
\newblock In \emph{Advances in Neural Information Processing Systems (NIPS)},
  pages 5279--5288, 2017.

\end{thebibliography}

\FloatBarrier

\section*{Appendix}

\appendix

\section{Solution for the Lagrange Multipliers}
\label{sec:lagrange_solution}
In order to compute a solution to the optimization objective with a KL-divergence and an entropy bound, we solve, using the dual of the problem, for the Lagrange multipliers associated with the bounds. We first discuss for the continuous action case how we optimize the multipliers exactly in the case of only log-linear parameters, continue with how we find an approximate solution in the case of also non-linear parameters, and then discuss the discrete action case.

\subsection{Computing Lagrange Multipliers $\eta$ and $\omega$ for log-linear Parameters}

Minimize the dual of the optimization objective (see e.g.\ \cite{akrour2016model-free} for a similar dual)
\begin{flalign}
  g_t(\eta, \omega) &= \eta \epsilon - \omega \beta + (\eta + \omega)
  \int \tilde{p}_t(\vec s) \log \left( \int
  \pi(\vec a| \vec s)^{\eta / (\eta + \omega)} \exp
  \left( \tilde{Q}_t(\vec s,\vec a) / (\eta + \omega) \right)
  d\vec a \right) d\vec s &
  \label{eq:g_eta_omega}
\end{flalign}
w.r.t.~$\eta$ and $\omega$. Note that action independent parts of $\log \pi(\vec a| \vec s)$ in Eq.~(\ref{eq:g_eta_omega}) do not have an effect
on the choice of $\eta$ and $\omega$ and we will discard them.

For a Gaussian policy 
\begin{equation}
  \pi(\vec a| \vec s) = \mathcal{N}\left(\vec a \Bigg| \vec \mu = \vec
  K \vec \varphi(\vec s) = \sum_i \varphi_i(\vec s) \vec k_i, \vec
  \Sigma \right)
\end{equation}
we get
\begin{flalign}
  \tilde{Q}_t(s,a) &= \vec \psi(\vec s, \vec a)^T \vec w =
  \begin{bmatrix}
  - \vecto[0.5  \vec a \vec a^T] \\
    \vecto[\vec a \vec \varphi(\vec s)^T]
  \end{bmatrix}^T 
  \begin{bmatrix}
    \vec w_1 \\
    \vec w_2
  \end{bmatrix}
  = - 0.5 \vec a^T \vec W_{aa} \vec a + \vec \varphi(\vec s)^T \vec W_{sa} \vec a &
\end{flalign}
\begin{flalign}
  g_t(\eta, \omega) &= \eta \epsilon - \omega \beta + (\eta + \omega)
  \int \tilde{p}_t(\vec s) \log \Bigg( \int C_{\pi}^{\frac{\eta}{\eta + \omega}}
  \exp \left(- 0.5 \eta / (\eta + \omega)
  (\vec a - \vec \mu)^T \vec \Sigma^{-1} (\vec a - \vec \mu) \right) &\nonumber\\
  & \exp \left(- 0.5 \vec a^T \vec W_{aa}/(\eta + \omega) \vec a 
               + \vec \varphi(\vec s)^T \vec W_{sa}/(\eta + \omega) \vec a \right)
    d\vec a \Bigg) d\vec s &
\end{flalign}
\begin{flalign}
  g_t(\eta, \omega) &= \eta \epsilon - \omega \beta + (\eta + \omega)
  \int \tilde{p}_t(\vec s) \log \Bigg( \int C_{\pi}^{\frac{\eta}{\eta + \omega}} &\nonumber\\
  &\quad \exp \left(\frac{\eta}{\eta + \omega}
    (- 0.5 \vec a^T \vec \Sigma^{-1} \vec a +
     \vec \mu^T \vec \Sigma^{-1} \vec a - 
     0.5 \vec \mu^T \vec \Sigma^{-1} \mu) \right) &\nonumber\\
  &\quad \exp \left(- 0.5 \vec a^T \vec W_{aa} / (\eta + \omega) \vec a +
                 \vec \varphi(\vec s)^T \vec W_{sa}/(\eta + \omega) \vec a \right)
    d\vec a \Bigg) d\vec s &
\end{flalign}
\begin{flalign}
  g_t(\eta, \omega) &= \eta \epsilon - \omega \beta + (\eta + \omega)
  \int \tilde{p}_t(\vec s) \log \Bigg( \int C_{\pi}^{\frac{\eta}{\eta + \omega}}
  \exp \Bigg( \frac{1}{\eta + \omega} \Bigg(
  - 0.5 \vec a^T \Big(\eta \vec \Sigma^{-1} +
                      \vec W_{aa} \Big) \vec a &\nonumber\\
  &\quad+ \Big(\eta
     \vec \varphi(\vec s)^T \vec K^T \vec \Sigma^{-1} +
     \vec \varphi(\vec s)^T \vec W_{sa} \Big) \vec a 
  - 0.5 \eta \vec \varphi(\vec s)^T \vec K^T \vec \Sigma^{-1} 
  \vec K \vec \varphi(\vec s) \Bigg) \Bigg)
  d\vec a \Bigg) d\vec s &
\end{flalign}
\begin{flalign}
  g_t(\eta, \omega) &= \eta \epsilon - \omega \beta + (\eta + \omega)
  \int \tilde{p}_t(\vec s) \log \Bigg( \int C_{\pi}^{\frac{\eta}{\eta + \omega}} &\nonumber\\
  &\quad \exp \left( \frac{1}{\eta + \omega} \left(
  - 0.5 \vec a^T \vec H_{aa} \vec a
  + \vec \varphi(\vec s)^T \vec H_{sa}\vec a
  - 0.5 \vec \varphi(\vec s)^T \vec H_{ss} \vec \varphi(\vec s) \right) \right)
  d\vec a \Bigg) d\vec s &
\end{flalign}
\begin{flalign}
  g_t(\eta, \omega) &= \eta \epsilon - \omega \beta + (\eta + \omega)
  \int \tilde{p}_t(\vec s) \log \Bigg( \int C_{\pi}^{\frac{\eta}{\eta + \omega}} &\nonumber\\
  &\quad \exp \Big( \frac{1}{\eta + \omega} \Big(
  - 0.5 (\vec a - \vec H_{aa}^{-1} \vec H_{sa}^T \vec \varphi(\vec s))^T
   \vec H_{aa} (\vec a - \vec H_{aa}^{-1} \vec H_{sa}^T \vec \varphi(\vec s)) &\nonumber\\
  &\quad\quad+ 0.5 \vec \varphi(\vec s)^T \vec H_{sa} \vec H_{aa}^{-1} \vec H_{sa}^T
        \vec \varphi(\vec s)
  - 0.5 \vec \varphi(\vec s)^T \vec H_{ss} \vec \varphi(\vec s) \Big) \Big)
  d\vec a \Bigg) d\vec s &
\end{flalign}
\begin{flalign}
  g_t(\eta, \omega) &= \eta \epsilon - \omega \beta + (\eta + \omega)
  \int \tilde{p}_t(\vec s) \log \Bigg(
  \exp \Big( \frac{1}{\eta + \omega} \Big(0.5 \vec \varphi(\vec s)^T (\vec H_{sa} \vec H_{aa}^{-1} \vec H_{sa}^T - \vec H_{ss}) \vec \varphi(\vec s)\Big) \Big) &\nonumber\\
  &((2\pi)^{-k / 2} |\vec \Sigma|^{-0.5})^{\frac{\eta}{\eta + \omega}} /
  ((2\pi)^{-k / 2} |(\eta + \omega) \vec H_{aa}^{-1}|^{-0.5}) \Bigg) d\vec s &
\end{flalign}
\begin{flalign}
  g_t(\eta, \omega) &= \eta \epsilon - \omega \beta + (\eta + \omega)
  \int \tilde{p}_t(\vec s)
  \Big(\frac{1}{\eta + \omega} 0.5 \vec \varphi(\vec s)^T (\vec H_{sa} \vec H_{aa}^{-1} 
  \vec H_{sa}^T - \vec H_{ss}) \vec \varphi(\vec s) &\nonumber\\
  &\quad+ \frac{\eta}{\eta + \omega} \log ((2\pi)^{-k / 2} |\vec \Sigma|^{-0.5})
   - \log((2\pi)^{-k / 2} |(\eta + \omega) \vec H_{aa}^{-1}|^{-0.5}) \Big) d\vec s, &
\end{flalign}
\begin{flalign}
  g_t(\eta, \omega) &= \eta \epsilon - \omega \beta +
  \int \tilde{p}_t(\vec s)
  \Big(0.5 \vec \varphi(\vec s)^T (\vec H_{sa} \vec H_{aa}^{-1} 
  \vec H_{sa}^T - \vec H_{ss}) \vec \varphi(\vec s) \Big) d\vec s &\nonumber\\
  &\quad+ \eta \log ((2\pi)^{-k / 2} |\vec \Sigma|^{-0.5})
   - (\eta + \omega) \log((2\pi)^{-k / 2} |(\eta + \omega) \vec H_{aa}^{-1}|^{-0.5}) , &
\end{flalign}
\begin{flalign}
  g_t(\eta, \omega) &= \eta \epsilon - \omega \beta +
     \int \tilde{p}_t(\vec s)
     \Big(0.5 \vec \varphi(\vec s)^T (\vec H_{sa} \vec H_{aa}^{-1} 
     \vec H_{sa}^T - \vec H_{ss}) \vec \varphi(\vec s) \Big) d\vec s &\nonumber\\
     &\quad- 0.5 \eta \log |2\pi \vec\Sigma|
      + 0.5 (\eta + \omega) \log |2\pi (\eta + \omega) \vec H_{aa}^{-1}| , &
\end{flalign}
where
\begin{flalign}
  \vec H_{aa} &= \eta \vec \Sigma^{-1} + \vec W_{aa} &\\
  \vec H_{sa} &= \eta \vec K^T \vec \Sigma^{-1} + \vec W_{sa} &\\
  \vec H_{ss} &= \eta \vec K^T \vec \Sigma^{-1} \vec K &\\
\end{flalign}
and $k$ is the dimensionality of actions. We got the end result by completing the square.

\subsection{Computing $\eta$ and $\omega$ for non-linear parameters}

Similarly to the log-linear parameters we minimize the dual
\begin{flalign}
  g_t(\eta, \omega) &= \eta \epsilon - \omega \beta + (\eta + \omega)
  \int \tilde{p}_t(\vec s) \log \left( \int
  \pi(\vec a| \vec s)^{\eta / (\eta + \omega)} \exp
  \left( \tilde{Q}_t(\vec s,\vec a) / (\eta + \omega) \right)
  d\vec a \right) d\vec s &
\end{flalign}
w.r.t.~$\eta$ and $\omega$. As before action independent parts of $\log \pi(\vec a| \vec s)$ in Eq.~(\ref{eq:g_eta_omega}) do not have an effect on the choice of $\eta$ and $\omega$ and we will discard them.

In our \emph{Linear Gaussian policy with constant covariance}
\begin{equation}
  \pi(\vec a| \vec s) = \mathcal{N}\left(\vec a \Bigg| \vec \mu = \vec
  K \vec \varphi(\vec s) = \sum_i \varphi_i(\vec s) \vec k_i, \vec
  \Sigma \right) ,
\end{equation}
we have
\begin{flalign}
  \log \pi(\vec a| \vec s) &= 
  - 0.5 \vec\varphi(\vec s)^T \vec U \vec K \vec\varphi(\vec s) 
  + \vec \varphi(\vec s)^T \vec U \vec a  
  - 0.5 \vec a^T \vec \Sigma^{-1} \vec a + \textrm{const},
\end{flalign}
where $\textrm{const} = - \sqrt{(2\pi)^k |\vec \Sigma|}$ and
$\vec U = \vec K^T \vec\Sigma^{-1}$. Therefore,
\begin{flalign}
  \nabla_{\vec \beta} \log \pi_{\vec \beta,\vec \theta}(\vec a|\vec s) &=
    \frac{\partial}{\partial \vec \beta}
  \vec \varphi(\vec s)^T \vec U \vec a
  -
  \frac{\partial}{\partial \vec \beta}
  0.5 \vec\varphi(\vec s)^T \vec U \vec K \vec\varphi(\vec s) ,
\end{flalign}
where we are able to split the
equation into action-value and value parts depending on whether they
depend on $\vec a$. Using the action-value part $\frac{\partial}{\partial
  \vec \beta} \vec \varphi(\vec s)^T \vec U \vec a$
to estimate $\vec w_3$ we get
\begin{flalign}
  \tilde{Q}_t(s,a) &= \vec \psi(\vec s, \vec a)^T \vec w =
  \begin{bmatrix}
  - \vecto[0.5  \vec a \vec a^T] \\
    \vecto[\vec a \vec \varphi(\vec s)^T] \\
    \frac{\partial}{\partial
      \vec \beta} \vec \varphi(\vec s)^T \vec U \vec a
  \end{bmatrix}^T 
  \begin{bmatrix}
    \vec w_1 \\
    \vec w_2 \\
    \vec w_3
  \end{bmatrix}
  = - 0.5 \vec a^T \vec W_{aa} \vec a + \vec \varphi(\vec s)^T \vec W_{sa} \vec a
  + \vec w_a(s)^T \vec a, &\\
\end{flalign}
where $\vec w_a(s)^T = \vec w_3^T \frac{\partial \vec \varphi(\vec
  s)^T}{\partial \vec \beta} \vec U$. By completing the square we get
\begin{flalign}
  g_t(\eta, \omega) &= \eta \epsilon - \omega \beta + (\eta + \omega)
    \int \tilde{p}_t(\vec s) \log \Bigg( \int C_{\pi}^{\frac{\eta}{\eta + \omega}}
    \exp \left(- 0.5 \eta / (\eta + \omega)
    (\vec a - \vec \mu)^T \vec \Sigma^{-1} (\vec a - \vec \mu) \right) &\nonumber\\
    & \exp \left(- 0.5 \vec a^T \vec W_{aa}/(\eta + \omega) \vec a 
                 + (\vec \varphi(\vec s)^T \vec W_{sa} + \vec
                   w_a(s)^T)/(\eta + \omega) \vec a \right)
      d\vec a \Bigg) d\vec s &
\end{flalign}
\begin{flalign}
   g_t(\eta, \omega) &= \eta \epsilon - \omega \beta + (\eta + \omega)
    \int \tilde{p}_t(\vec s) \log \Bigg( \int C_{\pi}^{\frac{\eta}{\eta + \omega}}
    \exp \Bigg( \frac{1}{\eta + \omega} \Bigg(
    - 0.5 \vec a^T \Big(\eta \vec \Sigma^{-1} +
                        \vec W_{aa} \Big) \vec a &\nonumber\\
    &\quad+ \Big(\eta
       \vec \varphi(\vec s)^T \vec K^T \vec \Sigma^{-1} +
       (\vec \varphi(\vec s)^T \vec W_{sa} + \vec
         w_a(s)^T) \Big) \vec a 
    - 0.5 \eta \vec \varphi(\vec s)^T \vec K^T \vec \Sigma^{-1} 
    \vec K \vec \varphi(\vec s) \Bigg) \Bigg)
    d\vec a \Bigg) d\vec s &
\end{flalign}
\begin{flalign}
 g_t(\eta, \omega) &= \eta \epsilon - \omega \beta +
  \int \tilde{p}_t(\vec s)
  \Big(0.5 (\vec h_{a}(\vec s)^T \vec H_{aa}^{-1} \vec h_{a}(\vec s) -
            h_{ss}(\vec s)) d\vec s &\nonumber\\
  &\quad- 0.5 \eta \log |2\pi \vec\Sigma|
      + 0.5 (\eta + \omega) \log |2\pi (\eta + \omega) \vec H_{aa}^{-1}| , &
\end{flalign}
where
\begin{flalign}
  \vec H_{aa} &= \eta \vec \Sigma^{-1} + \vec W_{aa} &\\
  \vec h_a(\vec s) &= \eta \vec \varphi(\vec s)^T \vec
  K^T \vec \Sigma^{-1} + \vec \varphi(\vec s)^T \vec W_{sa} + \vec
  w_a(s)^T &\\
  h_{ss}(\vec s) &= \eta \varphi(\vec s)^T \vec K^T \vec \Sigma^{-1} \vec K \vec \varphi(\vec s) &\\
\end{flalign}
and $k$ is the dimensionality of actions.

\subsection{Derivation of the dual for the discrete action case}
To derive the dual of our trust region optimization problem with entropy regularization and discrete actions we start with following program, where we replaced the expectation with integrals and use the compatible value function for the returns.
\begin{align}
\textrm{argmax}_{\pi_{\vec{\theta}}} 
& \; \int p(\vec{s}) \int \pi_{\vec{\theta}}(\vec{a}|\vec{s}) \; \tilde{G}^{\pi_\textrm{old}}_{\vec w}(\vec s, \vec a)  \; d\vec{a} \; d\vec{s} \\
\text{subject to } & \; \int p(\vec{s}) \int KL \left( \pi_{\vec{\theta}}(\vec{a}|  \vec{s}) \; || \; \pi_{\vec{\theta}_{\text{old}}}( \vec{a} | \vec{s}) \right)  \; d\vec{a} \; d\vec{s} < \epsilon \nonumber \\
& \; \int p(\vec{s}) \int H \left( \pi_{\vec{\theta}}(\vec{a}|\vec{s}) \right) - H \left( \pi_{\vec{\theta}_{\text{old}}}( \vec{a}|\vec{s}) \right) \; d\vec{a} \; d\vec{s} < \beta  \nonumber \\
& 1 = \int \int p(\vec{s}) \pi_{\vec{\theta}}(\vec{a}|\vec{s}) \; d\vec{a} \; d \vec{s} \nonumber
\end{align}
Since we are in the discrete action case we will be using sums for the brevity of the derivation, but the same derivation can be also done with integrals. Using the method of Lagrange multipliers \cite{boyd2004convex}, we obtain following Lagrange
\begin{align}
\label{eq:L_Lagrange} 
L(\pisa, \eta, \omega, \lambda) &= - \sum_{\vec{s},\vec{a}} p(\vec{s}) \pisa  \tilde{G}^{\pi_{old}}_{w}(\vec{a},\vec{s}) \\
&\quad + \; \eta \left[ \sum_{\vec{s}} p(\vec{s}) \sum_{\vec{a}} \pisa \log \left( \frac{\pisa}{\pioldsa} \right) - \epsilon \right] \nonumber \\ 
&\quad + \; \omega \left[ \sum_{\vec{s},\vec{a}} p(\vec{s}) H(\piold(\vec{a}|\vec{s})) +  \sum_{\vec{s},\vec{a}} p(\vec{s}) \pisa \log( \pisa) - \beta \right] \nonumber \\
&\quad + \; \lambda \left[ \sum_{\vec{s},\vec{a}} p(\vec{s}) \pisa - 1 \right] \nonumber \\ 
\end{align}
We differentiate now the Lagrange with respect to $\pisa$ and obtain following system
\begin{align}
\delta_{\pisa} L &= p(\vec{s}) \left[-\tilde{G}^{\piold}_{\vec{w}}(\vec{s}, \vec{a}) + \eta \log \left( \frac{\pisa}{\pioldsa} \right) + \eta + \omega \log(\pisa) + \omega + \lambda \right].
\end{align}
Setting it to zero and rearranging terms results in
\begin{align}
\pisa &= \pioldsa^{\frac{\eta}{\eta+\omega}} \exp \left( \frac{\tilde{G}^{\piold}_{\vec{w}}(\vec{s}, \vec{a})}{\eta+\omega} \right) \exp\left( \frac{- \eta - \omega - \lambda }{\eta+\omega} \right) \label{eq:psa_sol}
\end{align}
where the last term can be seen as a normalization constant
\begin{align}
\exp\left( \frac{- \eta - \omega - \lambda }{\eta+\omega} \right) &=  \left( \sum_{\vec{a}} \pioldsa^{\frac{\eta}{\eta+\omega}} \exp \left( \frac{\tilde{G}^{\piold}_{\vec{w}}(\vec{a},\vec{s})}{\eta+\omega} \right) \right)^{-1} \label{eq:lambda}.
\end{align}
Plugging Eq.~(\ref{eq:psa_sol}) and Eq.~(\ref{eq:lambda}) into Eq.~(\ref{eq:L_Lagrange}) results in the dual (similar to \cite{akrour2016model-free}) used for optimization
\begin{align*}
L(\eta, \omega) =& - \eta \epsilon - \omega \beta + \omega \sum_{\vec{s},\vec{a}} p(\vec{s}) H(\piold(\vec{a}|\vec{s})) &\\
&- (\eta+\omega) \sum_{\vec{s}} p(\vec{s}) \log \left( \sum_{\vec{a}} \pioldsa^{\frac{\eta}{\eta+\omega}} \exp \left( \frac{\tilde{G}^{\piold}_{\vec{w}}(\vec{a},\vec{s})}{\eta+\omega} \right) \right).& \end{align*}

\section{Technical Details for Discrete Action Experiments}
\label{sec:technical_details_for_experiments}

Here, we provide details on the experiments with discrete actions. Table~\ref{tab:fvrs_details} shows details on the hyper-parameters used in the Field Vision RockSample (FVRS) experiments and Algorithm~\ref{algo:COPSconjugateGradient}
describes details on the discrete action algorithm.

\begin{table}[h]
	\centering
  	\tabcolsep=0.05cm
	\begin{tabular}{c|cccccc}
		& $(5,5)$ full & $(5,5)$ noise & $(5,7)$ full & $(5,7)$ noise & $(7,8)$ full & $(7,8)$ noise  \\
		\hline
		Input space dim. & 15 & 85 & 17 & 115 & 22 & 134 \\ 
		Output space dim. & 5 & 5 & 5 & 5 & 5 &  5 \\
		\# policy parameters & 1410 & 3510 & 1650  & 4560 & 1620 & 4980 \\
		\hline
		Sim. step per iteration & 5000 & 5000 & 5000 & 5000 & 5000 & 5000\\
		Total num. of iterations & 600 & 600 & 600 & 600 & 600 & 600\\
		Discount $\gamma$ & 0.95 & 0.95 & 0.95 & 0.95 & 0.95 & 0.95\\
		History Length & 1 & 15 & 1 & 15 & 1 & 15\\
		Horizon & 25 & 25 & 35 & 35 & 50 & 50 \\
	\end{tabular}
	\caption{Parameters used for FVRS instances}
	\label{tab:fvrs_details}
\end{table}

\begin{algorithm}
	\begin{algorithmic}
		\State Initialize policy network $\pi_{\vec{\theta}, \vec{\beta}}$ with non-linear parameters $\vec{\beta}$ and linear parameters $\vec{\theta}$ and $\Theta = (\vec{\theta}, \vec{\beta})$
		\For{episode $\gets 1$ \textbf{to} maxEpisode}
		\State Initialize empty batch $\mathcal{B}$\;
		\While{collected samples $<$ batchsize}
		\State \parbox[t]{\dimexpr0.95\linewidth-\algorithmicindent}{Run policy $\pi_{\vec{\theta}, \vec{\beta}}(\vec a | \vec s)$  for $T$ timesteps or until termination: Draw action $\vec{a_t} \sim \pi_{\vec{\theta}, \vec{\beta}}(\vec{a_t}| \vec{s_t})$, observe reward $r_t$\strut}
		\State Add samples $(\vec{s_t}, \vec{a_t}, r_t)$ to $\mathcal{B}$
		\EndWhile
		\State Compute advantage values $A^{\pi_{\text{old}}}(\vec{s}_i, \vec{a}_i)$
		\State Compute $\vec{w}=(\vec{w}_{\vec{\theta}}, \vec{w}_{\vec{\beta}})$ using conjugate gradient to solve
		\begin{equation*}
		\vec w = \vec F^{-1} \nabla_{\Theta} J_{\text{PG}}(\pi_{\Theta}) \; |_{\Theta = \Theta_{\text{old}}} \qquad \nabla_{\Theta} J_{\text{PG}}(\pi_{\Theta}) = \sum_i^{|\mathcal{B}|} \nabla_{\Theta} \log \pi_{\Theta}(\vec{a}_i| \vec{s}_i)
		\; A^{\pi_{\text{old}}}(\vec{s}_i, \vec{a}_i)		\end{equation*}
		\State Use $\tilde{G}^{\pi_\textrm{old}}_{\vec w}(\vec s, \vec a)$ to solve for $\eta > 0$ and $\omega >0$ using the dual to the corresponding trust region optimization problem:
		\begin{align*}
		\textrm{argmax}_{\pi_{\vec{\theta}}} 
		& \; \mathbb{E}_{\vec{s} \sim p(\vec{s})} \left[ \int  \pi_{\vec{\theta}}(\vec{a}|\vec{s}) \; \tilde{G}^{\pi_\textrm{old}}_{\vec w}(\vec s, \vec a) \; d\vec{a} \right] \\
		\text{subject to } & \; \mathbb{E}_{\vec{s} \sim p(\vec{s})} \left[ KL \left( \pi_{\vec{\theta}}(\; \cdot \;| \; \vec{s}) \; || \; \pi_{\vec{\theta}_{\text{old}}}(\; \cdot \; | \; \vec{s}) \right) \right] < \epsilon \\
		& \; \mathbb{E}_{\vec{s} \sim p(\vec{s})} \left[ H \left( \pi_{\vec{\theta}}(\; \cdot \;| \; \vec{s}) \right) - H \left( \pi_{\vec{\theta}_{\text{old}}}(\; \cdot \; | \; \vec{s}) \right) \right] < \beta
		\end{align*}
		\State Apply updates for the new policy:
		\begin{equation*}
		\vec \theta_{\text{new}} = \frac{\eta \vec \theta_{\textrm{old}} + \vec w_{\vec \theta}}{\eta + \omega} \quad \vec \beta_{\text{new}} = \vec \beta_{old} + s \frac{\vec w_{\vec \beta}}{\eta }
		\end{equation*}
		\State where $s$ is a rescaling factor found by line search
		\State 
		\EndFor
	\end{algorithmic}
	\caption{COPOS discrete actions}\label{algo:COPSconjugateGradient}
\end{algorithm}

\section{Additional Continuous Control Experiments with a TRPO Entropy Bonus}
\label{sec:TRPO_entropy_bonus}

Table~\ref{tab:continuous_TRPO_bonus} shows additional results for continuous control in the Roboschool environment. In these experiments, an additonal entropy bonus is added to the objective function of TRPO.

\begin{table}[h]
  \centering
  \tabcolsep=0.055cm
  \caption{\label{tab:continuous_problems_trpo_bound} Additional continuous control environment
    benchmark runs. In these experiments TRPO was run with an additional entropy term multiplied with a factor $\beta$ added to the objective function. All algorithms are TRPO versions with the two different policy structures $\pi_1(a|s)$, $\pi_2(a|s)$ and different $\beta$ except for the two COPOS entries. We report the mean of
    the average return over $50$ last iterations $\pm$ standard error
    over $10$ random seeds. Bold denotes: no statistically significant
    difference to the best result (Welch's t-test with $p <
    0.05$).}
  \label{tab:continuous_TRPO_bonus}
  \begin{tabular}{lrrrr}
    \hline\noalign{\smallskip}
    & COPOS $\beta$=auto & COPOS $\beta=\epsilon$ &  $\pi_1, \beta=0.02$ &  $\pi_2, \beta=0.02$ \\	
    \noalign{\smallskip}\hline\noalign{\smallskip}
	RoboInvDblPendulum-v1 &  7110.0  $\pm$  416.0 &  7720.0  $\pm$  235.0 &  8010.0  $\pm$  58.0  &  6600.0  $\pm$  330.0 \\
RoboHopper-v1 &  \textbf{ 2510.0  $\pm$  35.1 } &  \textbf{ 2430.0  $\pm$  48.0 } &  1370.0  $\pm$  37.1 &  703.0  $\pm$  50.1 \\
RoboWalker2d-v1 &  \textbf{ 1880.0  $\pm$  34.6 } &  \textbf{ 1910.0  $\pm$  41.5 } &  886.0  $\pm$  33.3 &  69.3  $\pm$  1.83 \\
RoboHalfCheetah-v1 &  \textbf{ 2800.0  $\pm$  57.3 } &  \textbf{ 2770.0  $\pm$  42.2 } &  1500.0  $\pm$  20.9 &  376.0  $\pm$  42.6 \\
RoboAnt-v1 &  \textbf{ 2330.0  $\pm$  49.7 } &  \textbf{ 2380.0  $\pm$  26.9 } &  1190.0  $\pm$  41.0 &  614.0  $\pm$  25.0 \\
RoboHumanoid-v1 &  \textbf{ 114.0  $\pm$  0.937 } &  52.9  $\pm$  0.306 &  20.2  $\pm$  0.515 &  -7.29  $\pm$  2.0 \\
RoboHumanFlagrunHard-v1 &  \textbf{ 77.7  $\pm$  5.12 } &  32.1  $\pm$  1.34 &  -121.0  $\pm$  11.3 &  -55.4  $\pm$  1.49 \\
RoboAtlasForwardWalk-v1 &  \textbf{ 238.0  $\pm$  2.08 } &  187.0  $\pm$  1.18 &  108.0  $\pm$  0.825 &  73.9  $\pm$  1.04 \\
    \noalign{\smallskip}\hline
  \end{tabular}
  \begin{tabular}{lrrrr}
      \hline\noalign{\smallskip}
      &  $\pi_1, \beta=0.01$   &  $\pi_2, \beta=0.01$   &  $\pi_1, \beta=0.005$   &  $\pi_2, \beta=0.005$ \\
      \noalign{\smallskip}\hline\noalign{\smallskip}
	RoboInvDblPendulum-v1 &  8030.0  $\pm$  124.0 &  7630.0  $\pm$  172.0 &  7730.0  $\pm$  222.0  &  7780.0  $\pm$  257.0 \\
	RoboHopper-v1 &  1620.0  $\pm$  18.4 &  1530.0  $\pm$  71.2 &  1940.0  $\pm$  29.7 &  1930.0  $\pm$  54.0 \\
	RoboWalker2d-v1 &  1140.0  $\pm$  39.3 &  757.0  $\pm$  91.0 &  1430.0  $\pm$  25.5 &  939.0  $\pm$  96.3 \\
	RoboHalfCheetah-v1 &  1840.0  $\pm$  25.1 &  1250.0  $\pm$  63.0 &  2180.0  $\pm$  16.6 &  2190.0  $\pm$  67.7 \\
	RoboAnt-v1 &  1930.0  $\pm$  33.8 &  1910.0  $\pm$  40.9 &  2220.0  $\pm$  44.0 &  2130.0  $\pm$  57.8 \\
	RoboHumanoid-v1 &  46.6  $\pm$  2.03 &  56.5  $\pm$  2.31 &  59.7  $\pm$  3.99 &  95.3  $\pm$  3.52 \\
	RoboHumanFlagrunHard-v1 &  -58.1  $\pm$  2.37 &  -36.1  $\pm$  1.96 &  -23.4  $\pm$  1.93 &  -11.5  $\pm$  1.78 \\
	RoboAtlasForwardWalk-v1 &  122.0  $\pm$  0.692 &  87.2  $\pm$  1.32 &  150.0  $\pm$  0.538 &  136.0  $\pm$  1.72 \\
      \noalign{\smallskip}\hline
  \end{tabular}
  \begin{tabular}{lrrrr}
      \hline
      &  $\pi_1$ $\beta=-0.02$   &  $\pi_2, \beta=-0.02$   &  $\pi_1, \beta=-0.01$   &  $\pi_2, \beta=-0.01$ \\
      \noalign{\smallskip}\hline\noalign{\smallskip}
	RoboInvDblPendulum-v1 &  7760.0  $\pm$  208.0 &  7670.0  $\pm$  378.0 &  7340.0  $\pm$  335.0  &  7490.0  $\pm$  292.0 \\
	RoboHopper-v1 &  2200.0  $\pm$  31.6 &  1800.0  $\pm$  180.0 &  2130.0  $\pm$  36.8 &  1950.0  $\pm$  140.0 \\
	RoboWalker2d-v1 &  1340.0  $\pm$  148.0 &  653.0  $\pm$  47.5 &  1610.0  $\pm$  106.0 &  796.0  $\pm$  45.0 \\
	RoboHalfCheetah-v1 &  2520.0  $\pm$  20.9 &  1140.0  $\pm$  215.0 &  2470.0  $\pm$  40.0 &  1460.0  $\pm$  243.0 \\
	RoboAnt-v1 &  \textbf{ 2320.0  $\pm$  38.4 } &  1340.0  $\pm$  148.0 &  2270.0  $\pm$  34.4 &  1790.0  $\pm$  133.0 \\
	RoboHumanoid-v1 &  87.1  $\pm$  3.36 &  98.1  $\pm$  3.18 &  77.1  $\pm$  2.09 &  106.0  $\pm$  2.76 \\
	RoboHumanFlagrunHard-v1 &  \textbf{ 81.1  $\pm$  4.31 } &  36.8  $\pm$  1.8 &  56.3  $\pm$  1.5 &  37.1  $\pm$  2.07 \\
	RoboAtlasForwardWalk-v1 &  199.0  $\pm$  0.908 &  212.0  $\pm$  4.19 &  195.0  $\pm$  1.23 &  226.0  $\pm$  2.34 \\
      \noalign{\smallskip}\hline
  \end{tabular}
  \begin{tabular}{lrr}
      \hline
      &  $\pi_1, \beta=-0.005$   &  $\pi_2, \beta=-0.005$ \\
      \noalign{\smallskip}\hline\noalign{\smallskip}
  RoboInvDblPendulum-v1 &   7230.0  $\pm$  348.0 &   7630.0 $\pm$  265.0  \\
   RoboHopper-v1 &  2130.0  $\pm$  32.2 &  2040.0  $\pm$  146.0 \\
   RoboWalker2d-v1 &  1430.0  $\pm$  153.0 &  962.0  $\pm$  133.0 \\
   RoboHalfCheetah-v1 &  2410.0  $\pm$  17.0 &  1920.0  $\pm$  240.0 \\
   RoboAnt-v1 &  \textbf{ 2310.0  $\pm$  48.1 } &  2070.0  $\pm$  33.1 \\
   RoboHumanoid-v1 &  73.8  $\pm$  2.66 &  106.0  $\pm$  2.24 \\
   RoboHumanFlagrunHard-v1 &  44.2  $\pm$  1.85 &  27.4  $\pm$  2.86 \\
   RoboAtlasForwardWalk-v1 &  189.0  $\pm$  1.08 &  221.0  $\pm$  1.81 \\
      \noalign{\smallskip}\hline
  \end{tabular}
 
\end{table}

\end{document}